\pdfoutput=1
\documentclass[11pt]{article}
\usepackage[preprint]{acl}

\usepackage{iftex}
\ifPDFTeX\else
  \PackageWarningNoLine{camera_ready}{The 9-page fit is measured under pdfLaTeX; XeLaTeX and LuaLaTeX set this
    text about half a column longer. Compile with pdfLaTeX (on Overleaf: Menu > Compiler > pdfLaTeX)}
\fi
\usepackage{times}
\usepackage{latexsym}
\usepackage[T1]{fontenc}
\usepackage[utf8]{inputenc}
\usepackage{microtype}
\usepackage{inconsolata}

\usepackage{graphicx}
\usepackage{subcaption}
\usepackage{booktabs}
\usepackage{multirow}
\usepackage{tabularx}
\usepackage{array}
\usepackage{makecell}
\usepackage{float}
\usepackage{placeins}
\usepackage{amsmath}
\usepackage{amssymb}
\usepackage{xspace}
\usepackage{enumitem}
\usepackage{fontawesome5}
\usepackage[most]{tcolorbox}
\usepackage[capitalize,nameinlink,noabbrev]{cleveref}

\makeatletter
\def\section{\@startsection{section}{1}{\z@}{-1.6ex plus -0.4ex minus -.2ex}{1.0ex plus 0.2ex minus .1ex}%
  {\large\bfseries\raggedright}}
\def\subsection{\@startsection{subsection}{2}{\z@}{-1.4ex plus -0.4ex minus -.2ex}{0.6ex plus .2ex}%
  {\normalsize\bfseries\raggedright}}
\makeatother
\setlist{nosep}
\AtBeginDocument{\hypersetup{
  pdftitle={One Framework to Diagnose Them All: A Stratified Benchmark for Bengali LLM Hallucination},
  pdfauthor={Asif Shahriyar Sushmit et al.},
  pdfsubject={Findings of EACL 2027},
  pdfkeywords={hallucination detection, Bengali, low-resource languages, benchmark, LLM-as-a-judge}}}

\newcommand{\alik}{\textit{Alikbachan}\xspace}
\newcommand{\para}[1]{\vspace{1pt}\noindent\textbf{#1}\hspace{4pt}}
\newcommand{\field}[1]{\texttt{#1}}
\newcommand{\mode}[1]{\textit{#1}}
\newcommand{\task}[1]{\texttt{#1}}
\newcommand{\state}[1]{\textsc{#1}}

\newcommand{\ie}{\textit{i.e.}\xspace}
\newcommand{\vs}{\textit{vs.}\@\xspace}
\newcommand{\perse}{\textit{per se}\xspace}
\newcommand{\defacto}{\textit{de facto}\xspace}
\newcommand{\primafacie}{\textit{prima facie}\xspace}

\definecolor{ink}{HTML}{111827}
\definecolor{muted}{HTML}{5B6474}
\definecolor{indigo}{HTML}{211CBF}
\definecolor{hcyan}{HTML}{74D2FA}
\definecolor{violet}{HTML}{7A4DFF}
\definecolor{hpink}{HTML}{EC8DFF}
\definecolor{lavA}{HTML}{A7A8F0}
\definecolor{lavB}{HTML}{9194E4}
\definecolor{lavC}{HTML}{8180DE}
\definecolor{lavD}{HTML}{6E65CE}
\definecolor{hpurple}{HTML}{5445B5}
\definecolor{gain}{HTML}{1E8F7A}
\definecolor{loss}{HTML}{C2417A}
\definecolor{lavtint}{HTML}{ECECFC}
\definecolor{boxframe}{HTML}{A7A8F0}
\definecolor{boxtitle}{HTML}{8180DE}
\definecolor{boxback}{HTML}{F5F5FE}

\crefname{section}{\S}{\S\S}
\Crefname{section}{\S}{\S\S}
\crefname{appendix}{\S}{\S\S}
\Crefname{appendix}{\S}{\S\S}
\crefformat{section}{\S#2#1#3}
\crefformat{subsection}{\S#2#1#3}
\crefformat{appendix}{\S#2#1#3}
\crefformat{subappendix}{\S#2#1#3}
\Crefformat{section}{\S#2#1#3}
\Crefformat{subsection}{\S#2#1#3}
\Crefformat{appendix}{\S#2#1#3}
\Crefformat{subappendix}{\S#2#1#3}
\crefmultiformat{section}{\S#2#1#3}{ and~\S#2#1#3}{, \S#2#1#3}{ and~\S#2#1#3}
\crefmultiformat{subsection}{\S#2#1#3}{ and~\S#2#1#3}{, \S#2#1#3}{ and~\S#2#1#3}
\crefmultiformat{appendix}{\S#2#1#3}{ and~\S#2#1#3}{, \S#2#1#3}{ and~\S#2#1#3}
\crefmultiformat{subappendix}{\S#2#1#3}{ and~\S#2#1#3}{, \S#2#1#3}{ and~\S#2#1#3}
\Crefmultiformat{section}{\S#2#1#3}{ and~\S#2#1#3}{, \S#2#1#3}{ and~\S#2#1#3}
\Crefmultiformat{subsection}{\S#2#1#3}{ and~\S#2#1#3}{, \S#2#1#3}{ and~\S#2#1#3}
\Crefmultiformat{appendix}{\S#2#1#3}{ and~\S#2#1#3}{, \S#2#1#3}{ and~\S#2#1#3}
\Crefmultiformat{subappendix}{\S#2#1#3}{ and~\S#2#1#3}{, \S#2#1#3}{ and~\S#2#1#3}
\crefrangeformat{section}{\S#3#1#4--\S#5#2#6}
\crefrangeformat{appendix}{\S#3#1#4--\S#5#2#6}
\crefname{figure}{Figure}{Figures}
\crefname{table}{Table}{Tables}
\crefname{equation}{Eq.}{Eqs.}
\crefname{footnote}{footnote}{footnotes}

\makeatletter
\expandafter\def\csname bn@pg@name\endcsname{1}%
\expandafter\def\csname bn@dp@name\endcsname{0.451pt}
\expandafter\def\csname bn@pg@bidrohi\endcsname{2}%
\expandafter\def\csname bn@dp@bidrohi\endcsname{1.153pt}
\expandafter\def\csname bn@pg@samas\endcsname{3}%
\expandafter\def\csname bn@dp@samas\endcsname{0.322pt}
\expandafter\def\csname bn@pg@byasbakya\endcsname{4}%
\expandafter\def\csname bn@dp@byasbakya\endcsname{0.430pt}
\expandafter\def\csname bn@pg@guruchandali\endcsname{5}%
\expandafter\def\csname bn@dp@guruchandali\endcsname{0.332pt}
\expandafter\def\csname bn@pg@sadhu\endcsname{6}%
\expandafter\def\csname bn@dp@sadhu\endcsname{2.762pt}
\expandafter\def\csname bn@pg@cholito\endcsname{7}%
\expandafter\def\csname bn@dp@cholito\endcsname{0.332pt}
\expandafter\def\csname bn@pg@exprompt\endcsname{8}%
\expandafter\def\csname bn@dp@exprompt\endcsname{1.153pt}
\expandafter\def\csname bn@pg@exresponse\endcsname{9}%
\expandafter\def\csname bn@dp@exresponse\endcsname{2.546pt}

\DeclareRobustCommand{\bn}[1]{{%
  \edef\bn@scale{\strip@pt\dimexpr\f@size\p@/10\relax}%
  \raisebox{-\bn@scale\dimexpr\csname bn@dp@#1\endcsname\relax}{%
    \includegraphics[page=\csname bn@pg@#1\endcsname,scale=\bn@scale]{camera_ready/figures/bn/bn_snippets.pdf}}}}
\makeatother

\newlength{\cellwd}

\newcommand{\gainv}[1]{\textcolor{gain}{$+$#1}}
\newcommand{\lossv}[1]{\textcolor{loss}{$-$#1}}

\newtcolorbox{promptbox}[1]{enhanced, colback=boxback, colframe=boxframe, boxrule=0.6pt, arc=2pt,
  left=6pt, right=6pt, top=4pt, bottom=4pt, fonttitle=\bfseries\small, coltitle=white, colbacktitle=boxtitle,
  before title={\hypersetup{citecolor=white,linkcolor=white}}, title={#1}, fontupper=\small}

\providecommand{\teaservariant}{v1}
\providecommand{\pipelinevariant}{v1}
\newcommand{\figvariant}[2]{camera_ready/figures/variants/#1_#2.pdf}

\newcommand*{\repourl}{https://github.com/Alikbachan/Alikbachan}
\newcommand*{\kaggleurl}{https://www.kaggle.com/competitions/bengali-hallucination/}
\newcommand*{\hfurl}{https://huggingface.co/datasets/Alikbachan/Alikbachan}
\newcommand*{\hflabel}{Alikbachan/Alikbachan}   %
\newcommand*{\hficon}{\raisebox{-0.14em}{\includegraphics[height=1.0em]{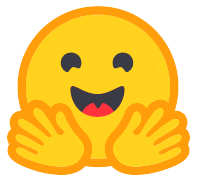}}}
\newcommand*{\projecturl}{https://alikbachan.github.io}
\newcommand*{\projectlabel}{alikbachan.github.io}
\definecolor{kaggleblue}{HTML}{20BEFF}   %
\newcommand{\linkentry}[3]{\ifx#2\empty\else\mbox{#1\;\href{#2}{\texttt{#3}}}\fi}
\newcommand{\linksep}[1]{\ifx#1\empty\else\hspace{1.25em}\fi}   %
\newcommand{\links}{{%
  \linkentry{\faGithub}{\repourl}{Alikbachan/Alikbachan}%
  \linksep{\kaggleurl}\linkentry{{\color{kaggleblue}\faKaggle}}{\kaggleurl}{bengali-hallucination}%
  \linksep{\hfurl}\linkentry{\hficon}{\hfurl}{\hflabel}%
  \linksep{\projecturl}\linkentry{\faGlobeAsia}{\projecturl}{\projectlabel}}}

\title{One Framework to Diagnose Them All:\\A Stratified Benchmark for Bengali LLM Hallucination}
\newcommand{\am}[1]{\textsuperscript{#1}}
\newcommand{\eqc}{$^{*}$}
\newcommand{\cor}{$^{\dagger}$}
\author{%
\begin{minipage}[t]{\textwidth}\centering\fontsize{10}{12.4}\selectfont\bfseries
Asif Shahriyar Sushmit\am{1}\eqc,\enspace
Shefayat E Shams Adib\am{2}\eqc,\enspace
Abdullah Al Jubaer Gem\am{2}\eqc,\\
Ahmed Shafin Ruhan\am{2}\eqc,\enspace
MD.\ Nazmuddoha Ansary\am{3}\eqc,\enspace
Biplab Kumar Sarkar\am{4}\eqc,\\
Syed Rifat Raiyan\am{2}\eqc\cor,\enspace
Ajwad Abrar\am{2},\enspace
Abdullah Bin Hasan\am{5},\enspace
Adib Sakhawat\am{2},\enspace
M.\ Sakeef Hossain\am{2},\\
Ahmed Alfey Sani\am{2},\enspace
Arefin Mizan\am{6},\enspace
Sourab Chakma\am{7},\enspace
Quazi Adibur Rahman Adib\am{8},\\
Wassie Haque\am{5},\enspace
Sanjay Kumar Sen\am{5},\enspace
Saad Tahmid\am{5},\enspace
Mohammad Mamun Or Rashid\am{9}\\[5pt]
\mdseries\small
\am{1}IEEE\quad
\am{2}Islamic University of Technology\quad
\am{3}Bengali.AI\quad
\am{4}North Dakota State University\\
\am{5}Bangladesh University of Engineering and Technology\quad
\am{6}Institute of Policy Dynamics\\
\am{7}Mawlana Bhashani Science and Technology University\quad
\am{8}BRAC University\quad
\am{9}Jahangirnagar University\\[4pt]
\links
\end{minipage}}

\newcommand{\authornotes}{%
  \begingroup
  \renewcommand{\thefootnote}{\fnsymbol{footnote}}%
  \footnotetext[1]{Equal contribution.}%
  \footnotetext[2]{Corresponding author: \href{mailto:rifatraiyan@iut-dhaka.edu}{\texttt{rifatraiyan@iut-dhaka.edu}}.}%
  \endgroup
  \setcounter{footnote}{0}}

\begin{document}
\maketitle
\authornotes
\begin{abstract}
Hallucination research in low-resource languages fragments easily: each newly named failure mode prompts its own
dataset, spending scarce annotators on incomparable resources. We argue for consolidation and test it at community
scale. \alik is a Bengali hallucination-detection benchmark stratified along three
crossed axes: a human-validated cultural-distance band, six task types, and a five-family, 32-mode hallucination
taxonomy. It comprises a harvested 2,516-item test set (Phase~1) and a quota-constructed, gold-labeled 5,000-item set
(Phase~2). The stratification is diagnostic. Four zero-shot LLM judges vary by over 30 macro-F1 points across strata,
and the item-level predictions of 99 Phase-1 and 32 Phase-2 competition systems reproduce the profile:
discourse-level omission defeats judges and most teams alike, whereas only a minority over-flag negative controls. The judges' verdict probabilities rank items well (ROC-AUC 0.74--0.95) but are overconfident (ECE
0.04--0.29). An open competition on \alik drew 222 teams and 3,566 submissions, and a single-prompt 32B judge reaches
the 57th percentile. Of the 25 teams in both phases, 24 regress on Phase~2 (mean $-17.6$ points) while every zero-shot
judge improves, and Phase-1 rank barely predicts Phase-2 rank (Spearman $\rho=0.35$), so a Phase-1 leaderboard alone
would have crowned a different winner.
\end{abstract}

\begin{figure}[!t]
  \centering
  \includegraphics[width=\columnwidth]{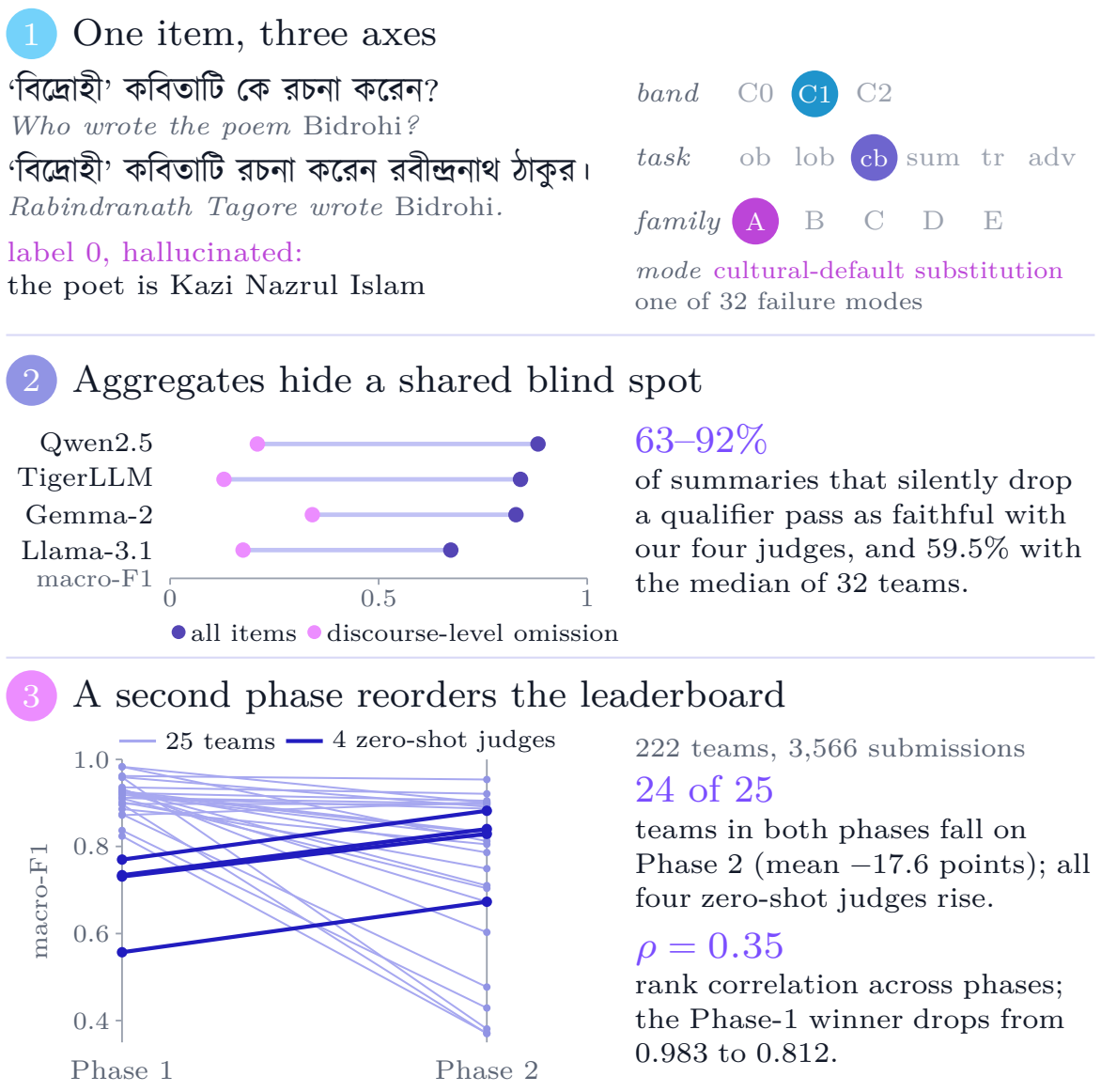}
  \caption{\alik at a glance. (1)~A Phase-2 item: asked who wrote the poem \textit{Bidrohi}, the response names
  Rabindranath Tagore instead of its author, Kazi Nazrul Islam; the item lies in band C1, closed-book QA and
  Family~A (cultural-default substitution), and its gold label is 0, hallucinated. (2)~Each zero-shot judge's macro-F1 on all of Phase~2 (dark dots,
  0.67--0.88) and on discourse-level omission (pink dots, 0.13--0.34): the four judges share a blind spot that their
  aggregate scores hide (\cref{sec:diagnosis}). (3)~Macro-F1 of the 25 teams in both phases (thin lines) and of the
  four judges (thick lines): from Phase~1 to Phase~2, 24 of the teams lose ground while every judge gains
  (\cref{sec:comp-experiments}).}
  \label{fig:teaser}
\end{figure}

\section{Introduction}
\label{sec:intro}

\begin{figure*}[t]
  \centering
  \includegraphics[width=\textwidth]{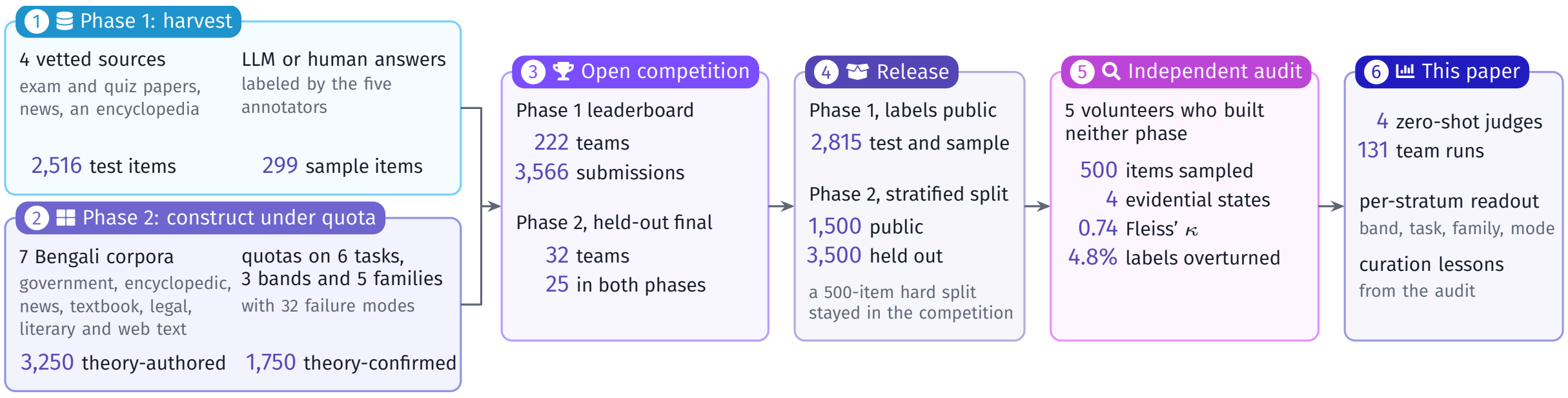}
  \caption{The \alik framework. Two independently sourced phases (1, 2) feed an open competition (3) and a public
  release (4). An independent audit (5) reviewed a sample of the released items, and this paper (6) analyzes zero-shot
  judges, the teams' item-level predictions and the audit (\cref{sec:dataset,sec:curation,sec:experiments}).}
  \label{fig:pipeline}
\end{figure*}

In a low-resource language, a fluent answer is both the hardest kind to doubt and the most expensive kind to check.
A model that writes grammatical, register-appropriate Bengali gives no sign when it misattributes a poem or misdates
a war, and few people can verify such claims against primary sources. The \defacto response to each newly
characterized failure is a new benchmark, a practice whose tacit premise is that such experts are plentiful. Most of
the world's languages lack them \citep{joshi2020state}. Spent one dataset at a time, a small pool of experts yields
isolated resources with incompatible schemas, labels and protocols, and claims about overall progress grow harder to
trust \citep{dehghani2021benchmarklottery,raji2021everything}.

Hallucination makes the cost palpable.\footnote{The word entered English in the 1640s from the Latin
\textit{hallucinationem}, `to wander in the mind'; computer vision once used it approvingly for algorithms that
envision missing pixels, before LLMs gave it the sense of fluent but fabricated text \citep{maleki2024ai}.} It is a
family of failures rather than one, and the family keeps growing \citep{ji2023survey}: detection benchmarks now
target parametric error, contextual infidelity, adversarial manipulation and calibration separately. Under
annotation scarcity, enumerating them one dataset at a time is a queue that never clears. The failures also interact
with language and culture in ways an English-first taxonomy does not anticipate. Factuality evaluation itself grows
less reliable as a language's resources shrink \citep{kim2024analysis}. Asked in Bengali who wrote the poem
\textit{Bidrohi}, a model may name Rabindranath Tagore, the Bengali poet best known abroad, instead of its author,
Kazi Nazrul Islam (\cref{fig:teaser}). Such an answer is grammatical and culturally legible, so nothing on its surface
betrays it.

We argue for consolidation over enumeration. If one resource is stratified along dimensions that cut across failure
modes, a newly identified phenomenon enters as another cell rather than another dataset. Benchmark and diagnostic
then dovetail: because every item carries its category metadata, one evaluation run reports where a system fails as
well as how well it scores. The axes themselves are not specific to Bengali. Cultural distance from a model's
dominant pretraining distribution, task type and an explicit hallucination taxonomy could serve wherever annotators
are scarce, although the bands and corpora must be instantiated locally.

Bengali is a germane test case: spoken by over 280 million people, it is resourced far below what that population
would predict. \alik (\bn{name}, `mirage of language') crosses three axes: a human-validated, three-level
cultural-distance band, six task types and a five-family, 32-mode hallucination taxonomy. Its items span general
knowledge, mathematics, Bengali grammar and lexicon, history, literature, law and current affairs, in two
independently sourced phases
(\cref{fig:pipeline}): a harvested Phase-1 test set of 2,516 items and a quota-constructed, gold-labeled Phase-2 set
of 5,000 items, of which 1,500 are public.

A resource's authors cannot settle the consolidation argument with their own evaluations, so we test it externally.
\alik is the foundation of an open research competition on Kaggle, with a public leaderboard phase followed by an
in-person final for the top teams. Its complete submission history is a litmus test for the argument: it shows
whether one stratified resource concentrates effort that would otherwise fragment, and whether a second stratified
phase catches what a leaderboard alone cannot. We make four contributions.
\begin{enumerate}[leftmargin=*,label=(\arabic*)]
  \item We design a stratification for annotator-constrained languages that crosses cultural distance, task type
  and hallucination family, and we document how both phases were built, labeled and independently audited
  (\cref{sec:dataset}).
  \item We show that the stratification is diagnostic. A judge's macro-F1 varies by more than 30 points across
  strata, and the item-level predictions of 99 Phase-1 and 32 Phase-2 competition systems reproduce the profile,
  including a blind spot for discourse-level omission that judges and most teams share (\cref{sec:experiments}).
  \item We analyze the complete submission history of a 222-team competition and, through the 25 teams that entered
  both phases, expose a generalization gap that neither phase reveals alone (\cref{sec:comp-experiments}).
  \item We distill an independent audit into curation lessons for low-resource benchmarks (\cref{sec:curation}).
\end{enumerate}

\section{Related Work}
\label{sec:related}

\para{Hallucination detection.} Detectors fall into three broad families: consistency-based methods that sample
several generations and flag disagreement \citep{manakul2023selfcheckgpt}; evidence-grounded methods that check a
response against a reference through entailment or question answering \citep{fabbri2022qafacteval,min2023factscore};
and supervised classifiers trained on labeled faithful and hallucinated pairs \citep{li2023halueval}. LLMs also
serve directly as judges \citep{zheng2023judging}, as our baselines do; we additionally measure how well calibrated
their verdicts are, a property long studied for neural classifiers \citep{guo2017calibration}. \alik serves all these
families (\cref{sec:baselines}).

\para{Multilingual and low-resource factuality.} Beyond the English-first benchmarks, Bengali now has a multi-task
hallucination evaluation \citep{adib2026benhalluevalmultitaskhallucinationevaluation} and a culturally grounded study
of idiom understanding \citep{sakhawat2026idioms}, but no stratified framework that diagnoses many hallucination types
at once. \citet{hu2020xtreme} showed that one unified multilingual benchmark can replace disjoint
monolingual ones; we apply the same principle within a single low-resource language.

\para{Benchmark proliferation.} Dehghani et al.\ and Raji et al.\ argue that a sprawling, poorly coordinated set of
benchmarks makes claims about progress hard to trust, and that effort is better spent on fewer, carefully designed
evaluations, an argument \cref{sec:consolidation} applies to low-resource hallucination detection. We know of no prior work that tests this argument against a full open
research competition rather than an internal set of baselines.

\para{Bengali resources and models.} Bengali benchmarks cover sentiment analysis \citep{kabir-etal-2023-banglabook},
medical named-entity recognition \citep{khan2023nervous}, social bias and cultural reasoning \citep{tasnim2026basco}
and text detoxification \citep{mohsin2025banglanirtoxlargescaleparallelcorpus}. Encoders such as BanglaBERT
\citep{bhattacharjee2022banglabert} and multilingual ones \citep{devlin2019bert,conneau2020xlmr} could anchor
supervised detectors. Bengali-adapted LLMs include TigerLLM \citep{raihan2025tigerllm}, TituLLM
\citep{nahin2025titullms} and BanglaLlama \citep{zehady2026banglallama}; we test all three as zero-shot judges.

\section{The Case Against Fragmentation}
\label{sec:consolidation}

A newly identified failure, such as a model substituting a Western officeholder for a Bangladeshi one, usually
prompts a targeted dataset. In Bengali this is costly. A small pool of experts must verify every item against
government documents, statutes, curricula and literary or historical sources, and single-purpose benchmarks drawn from
it duplicate effort and yield hard-to-compare results. We propose to spend the same effort on one framework instead
and state the desiderata it must meet.
\begin{enumerate}[leftmargin=*,label=(\roman*)]
  \item Axes that cut across failure modes, so that a new failure enters as a new combination of labels rather than
  a new dataset. \alik uses cultural distance, task type and a hallucination taxonomy (\cref{sec:bands,sec:tasktypes}).
  \item Item-level metadata, so that one evaluation run reports where a system fails as well as how well it scores
  (\cref{sec:diagnosis}).
  \item A second, independently constructed phase, so that overfitting to a public leaderboard becomes visible
  (\cref{sec:comp-experiments}).
  \item An independent audit that separates construction defects from model errors (\cref{sec:curation}).
\end{enumerate}
Internal evaluation alone cannot show that such a framework works at scale. If one well-stratified resource can
absorb broad participation without fragmenting into task-specific benchmarks, its value should persist across
independently built systems, and the open competition in \cref{sec:comp-experiments} tests exactly that.

\section{The \alik Framework}
\label{sec:dataset}

\subsection{Task Definition}
\label{sec:task}

Each item is a triple $(c,p,r)$ of an optional context $c$ (a passage, or the token \field{[NULL]}), a Bengali prompt
$p$ and a candidate response $r$. The task is to predict $y\in\{0,1\}$, where $y=1$ marks a faithful response and
$y=0$ a hallucinated one, \ie one with fabricated, unsupported or factually incorrect content. With a null context the
response is judged against world knowledge; otherwise it is judged against the context, even where the context
deliberately departs from reality. Both phases share this formulation and differ only in how their items were built
and in the metadata that accompany them. Every term used below is defined in \cref{app:glossary}.

\subsection{Two Phases, One Schema}
\label{sec:sources}

\alik was built along two independently sourced tracks (\cref{fig:pipeline}). \para{Phase~1} harvests short-answer
questions from four vetted Bengali sources (two examination and quiz paper collections, a newspaper corpus and an
encyclopedia) and pairs them with responses written by LLMs or people, so it follows a real question distribution
rather than a synthetic one. Of its 2,516 test items, 1,361 carry a supporting context and 1,155 are closed-book; a
further 299 labeled items form a sample set.

\para{Phase~2} is constructed under explicit quotas over six task types, three cultural bands and the 32 failure
modes of five families (\cref{sec:bands,sec:tasktypes}). Its items draw on seven Bengali corpora: Wikipedia and
Banglapedia (1,100 items), government publications (900), newspapers, National Curriculum and Textbook Board texts and
Bangladeshi statutes (700 each), e-Bangla Library texts (550) and Common Crawl (350). Of its 5,000 items, 3,250 are
theory-authored around a target failure mode and 1,750 are theory-confirmed, built from real source content. Every
quota was met exactly (\cref{app:p2-quota} gives the specification and the joint counts).

\para{Release.} During the competition the Phase-1 test labels sat behind the leaderboard; they are now public, as is
the sample set. Phase~2 is split 30/70 into a public set of 1,500 items, sampled so that every failure mode keeps its
share to within four items, and 3,500 items held out as a contamination-resistant test set. A further 500-item
hard split was used only in the competition. All analyses below use the full 5,000 Phase-2 items. The data are
released under CC BY-NC-SA 4.0, together with the judge scripts and a scorer that reports macro-F1, error rates,
ROC-AUC and ECE per stratum.

\subsection{Cultural-Distance Bands}
\label{sec:bands}

The central organizing axis is a three-level cultural-distance band. C0 (\textit{globally stable}) covers
culture-independent facts in science, geography or mathematics, such as the boiling point of water, and serves as the
control band. C1 (\textit{culturally situated}) covers answers that are specific to Bangladesh or differ from a
globally dominant default, such as Bengali literary authorship, Liberation War dates, national monuments and regional
administration; it hosts the headline phenomenon of fluent but locally wrong output. C2 (\textit{contested or
time-sensitive}) covers current officeholders and revised statistics, where a hallucination must be told apart from
outdated knowledge. Phase~2 fixes the band sizes at 750, 3,500 and 750 items. In Phase~1 the bands were assigned after
harvesting through a human-validated protocol (2,014, 483 and 19 items; \cref{app:phase1-detail}).

\subsection{Task Types and Hallucination Families}
\label{sec:tasktypes}

Phase~2 spans six task types: short- and long-form open-book QA (\task{short-ob-qa}, \task{long-ob-qa}), both
context-grounded; short-form closed-book QA (\task{short-cb-qa}); summarization; Bengali--English translation; and
adversarial probes built on false premises, prompt injections, counterfactual contexts or instruction constraints
(\task{adversarial}). Phase~1 distinguishes only closed-book from context-grounded QA, and with no separate
translation or long-form clusters it is best read as a short-answer QA benchmark.

Every Phase-2 item also belongs to one of five hallucination families, the diagnostic axis of
\cref{sec:experiments}: (A)~\textit{content or parametric} errors in dates, entities or attributions, independent of
any context; (B)~\textit{context or intrinsic} errors that contradict or embellish the given context;
(C)~\textit{linguistic or cultural} errors of Bengali script, register, lexicon or translation;
(D)~\textit{adversarial or robustness} failures, from prompt injection to miscalibration; and (E)~\textit{meta or
pipeline} artifacts, such as ambiguous gold labels, imperfect sources and encoding variants, that a robust judge
should see through. The families divide into 32 failure modes of 50--400 items each.

\subsection{Annotation and Audit}
\label{sec:annotation}

Two separate teams of five undergraduate volunteers built and checked \alik. The \textit{annotation team} labeled
every Phase-1 item and wrote Phase~2: theory-authored items are labeled by construction and theory-confirmed items by
review against their source. A 30-item spot-check of theory-authored items during construction corrected one label
(3.3\%).

The \textit{audit team}, fluent Bengali speakers, none of whom helped build either phase, reviewed the data after
both phases were complete. They were first trained on a four-way evidential protocol (\cref{sec:curation}) and given
few-shot examples of the three bands. They then reviewed a random sample of 500 items, 200 from the Phase-1 sample set
and 300 from Phase~2. Three auditors labeled a shared 100-item subset independently (Fleiss' $\kappa=0.74$) and split
the remaining 400 (about 133 each); disagreements, mostly between \state{Refuted} and \state{Not Entailed}, were
settled in a consensus discussion led by a senior researcher. The audit found 310 \state{Supported}, 65
\state{Refuted}, 100 \state{Not Entailed} and 25 \state{Malformed} items and overturned 24 of the 500 evidential
labels (4.8\%). A separate review of 250 items' band and domain assignments made 11 band and 8 domain changes.

\para{A worked example.} The first panel of \cref{fig:teaser} shows a theory-authored Phase-2 item from the Wikipedia
and Banglapedia track. Asked who wrote the poem \bn{bidrohi} (\textit{Bidrohi}, `The Rebel'), the response names
Rabindranath Tagore, the most widely known Bengali poet, instead of its author, Kazi Nazrul Islam. The answer depends on
Bengali literary history (band C1), the item has no context (\task{short-cb-qa}), and the error swaps a local fact for
a global default (Family~A, \mode{cultural-default substitution}). Its gold label is 0, and its \field{rationale} field
states why; the full record is printed in \cref{app:glossary}.

\section{Curation Lessons: A Cautionary Tale}
\label{sec:curation}

A binary faithful/hallucinated label suits scoring but collapses four evidential states: \state{Supported} (the
evidence justifies the response), \state{Refuted} (a necessary claim contradicts it), \state{Not Entailed} (required
support is absent) and \state{Malformed} (missing or corrupted content prevents judgment). The audit
(\cref{sec:annotation}) recorded the four-way state before projecting it onto the binary label and reviewed the band
and domain assignments. Its lessons reach beyond \alik.

\begin{table*}[t]
\centering
\small
\setlength{\tabcolsep}{5pt}
\begin{tabularx}{\textwidth}{@{}>{\raggedright\arraybackslash}p{0.125\textwidth}>{\raggedright\arraybackslash}p{0.405\textwidth}>{\raggedright\arraybackslash}X@{}}
\toprule
\textbf{Layer} & \textbf{Common pitfall} & \textbf{Required curator check} \\
\midrule
Packaging & An image, option, table or answer-bearing part is missing from the released row & Test answerability on the released row itself, not on its source \\
Representation & OCR, Unicode or math notation admits competing readings & Keep raw and canonical forms; adjudicate only once one reading is fixed \\
Linguistic validity & The idiom, register, morphology or lexical relation is ill-formed & Judge grammaticality, lexical correctness and contextual support separately \\
Question contract & The answer has the wrong role, entity, place, time, unit or polarity & Record an answer-slot signature that binds every requested argument \\
Epistemic status & An estimate is treated as exact, or an unresolved future as fact & Preserve approximation, modality, authority and temporal scope \\
Reasoning & Topical overlap stands in for a comparison or composition & Require a licensed operation and reject the nearest plausible alternative \\
Annotation record & Response, rationale and label describe different judgments & Reconcile every field before release; one matching field cannot repair another \\
\bottomrule
\end{tabularx}

\caption{The independent audit condensed into a seven-layer item contract: for each layer, the pitfall the audit found
in released items and the check a curator should run. Layers are ordered by where adjudication should stop, since an
item that fails one layer cannot be judged on the layers below it; \cref{app:curation-detail} gives the full tables.}
\label{tab:audit-contract}
\end{table*}

\para{Bands define the verification an item needs.} A C0 item can still be malformed by wrong notation, units or
polarity; a culturally plausible C1 response can attach a correct fact to the wrong entity; and two C2 values can both
be right under different reference dates. Band membership therefore says how an item must be checked, and passing
that check is a separate matter. \Cref{tab:audit-contract} condenses the audit into a seven-layer contract.

\para{Most contextual errors break a relation.} The commonest contextual failures reuse the context's vocabulary but
assign it to the wrong entity, role, date, jurisdiction or comparison: role substitution, jurisdiction transfer,
near-date selection. Lexical overlap cannot catch them. The audit instead records an answer-slot signature, a
statement of what a correct answer must be (its type, entity, relation, jurisdiction, time, ordinal, polarity, unit
and precision), which the response and its supporting span must satisfy together.

\para{Some apparent hallucinations are construction defects.} Input-side defects, such as a multiple-choice stem that
refers to dropped options or an ambiguous OCR reading, can make a faithful response look hallucinated. So can source
or time defects, such as a copied answer key with an upstream error, or a stale snapshot. Such items belong in a
repair queue before evaluation.

\para{Linguistic validity belongs in the taxonomy.} Antonymy, synonymy, idiom interpretation, register and
compounding are distinct operations that items can conflate, and the audit found topically related terms wrongly
accepted as satisfying a lexical relation. We folded these cases into Family~C rather than a separate resource, so a
newly found failure became new taxonomy cells, as \cref{sec:consolidation} anticipates. The audit further recommends
keeping clean and perturbed variants, recording the four-way state before the binary projection, and asking
adjudicators to state why the nearest alternative fails; these practices carry over to other low-resource languages.

\section{Diagnosis in Practice}
\label{sec:experiments}

We report results on the two canonical evaluations, the Phase-1 test set (2,516 items) and the full Phase-2 set
(5,000 items), so that every number is comparable across analyses.

\subsection{Four Zero-Shot Judges}
\label{sec:baselines}

We evaluate four zero-shot LLM judges spanning about an order of magnitude in scale: Qwen2.5-32B
\citep{qwen2025qwen25}, Gemma-2-9B \citep{gemmateam2024gemma2}, Llama-3.1-8B \citep{grattafiori2024llama3} and
TigerLLM-9B-it, the Bengali-adapted Gemma-2-9B. All four receive the same prompt (\cref{app:prompt}), a strict
instruction to output one verdict digit in the dataset's convention (1 faithful, 0 hallucinated), and decode greedily.
Qwen2.5-32B, Gemma-2-9B and TigerLLM-9B-it run locally in bfloat16, and Llama-3.1-8B through an API. The smaller Bengali-adapted TituLLM-1B and BanglaLlama-7B stayed near the majority-class reference
and were dropped (\cref{app:bengali-llms}).

From the log-probabilities of the two verdict tokens we compute $s=P(1)/(P(1)+P(0))$, the judge's probability that
the response is faithful. Besides accuracy, macro-F1 and per-class F1, \cref{tab:headline} reports threshold-free
PR-AUC for each class and ROC-AUC, which matter under class imbalance \citep{vashurin2025benchmarking}, and the
expected calibration error (ECE) of the confidence $\max(s,1-s)$.

\begin{table*}[t]
\centering
\small
\setlength{\tabcolsep}{4.4pt}
\begin{tabular}{@{}lrrrrrrrr@{}}
\toprule
 & \multicolumn{4}{c}{\textbf{Verdicts}} & \multicolumn{4}{c}{\textbf{Verdict probabilities}} \\
\cmidrule(lr){2-5}\cmidrule(l){6-9}
\textbf{System} & \textbf{Acc.} & \textbf{Macro-F1} & \textbf{F1}$_{\text{hallu}}$ & \textbf{F1}$_{\text{faith}}$ & \textbf{PR-AUC}$_{\text{faith}}$ & \textbf{PR-AUC}$_{\text{hallu}}$ & \textbf{ROC-AUC} & \textbf{ECE}\,$\downarrow$ \\
\midrule
\multicolumn{9}{@{}l}{\textit{Phase-1 test set (2{,}516 items)}} \\
Majority class & 0.531 & 0.347 & 0.000 & 0.693 & 0.531 & 0.469 & 0.500 & -- \\
Qwen2.5-32B & \textbf{0.770} & \textbf{0.770} & \textbf{0.781} & 0.758 & \textbf{0.894} & \textbf{0.851} & \textbf{0.870} & 0.212 \\
Gemma-2-9B & 0.733 & 0.731 & 0.711 & 0.751 & 0.851 & 0.810 & 0.827 & 0.227 \\
Llama-3.1-8B & 0.631 & 0.557 & 0.376 & 0.738 & 0.810 & 0.761 & 0.776 & 0.285 \\
TigerLLM-9B-it & 0.741 & 0.734 & 0.694 & \textbf{0.775} & 0.871 & 0.832 & 0.847 & \textbf{0.061} \\
\midrule
\multicolumn{9}{@{}l}{\textit{Phase~2 (5{,}000 items)}} \\
Majority class & 0.543 & 0.352 & 0.704 & 0.000 & 0.457 & 0.543 & 0.500 & -- \\
Qwen2.5-32B & \textbf{0.883} & \textbf{0.882} & \textbf{0.893} & \textbf{0.871} & \textbf{0.936} & \textbf{0.962} & \textbf{0.951} & 0.108 \\
Gemma-2-9B & 0.829 & 0.829 & 0.834 & 0.824 & 0.873 & 0.934 & 0.910 & 0.147 \\
Llama-3.1-8B & 0.682 & 0.673 & 0.619 & 0.727 & 0.627 & 0.817 & 0.741 & 0.206 \\
TigerLLM-9B-it & 0.840 & 0.840 & 0.836 & 0.844 & 0.887 & 0.937 & 0.919 & \textbf{0.042} \\
\bottomrule
\end{tabular}

\caption{Zero-shot judges against a majority-class reference on both releases. The verdict columns score the 0/1
verdicts; the probability columns score $s=P(1)/(P(1)+P(0))$, with PR-AUC taking the faithful or the hallucinated class
as positive and ECE (lower is better) binning the confidence $\max(s,1-s)$ into 15 equal-width bins. For the majority
reference PR-AUC equals the positive-class prevalence and ROC-AUC is 0.5. Bold marks the best judge per column:
Qwen2.5-32B leads on both releases, and TigerLLM-9B-it is the best calibrated.}
\label{tab:headline}
\end{table*}

Qwen2.5-32B is the strongest judge on both releases (macro-F1 0.770 and 0.882), followed by TigerLLM-9B-it (0.734,
0.840) and Gemma-2-9B (0.731, 0.829); Llama-3.1-8B trails (0.557, 0.673) but stays above the majority reference.
PR-AUC keeps this order and places every judge above the prevalence baseline for both classes (on Phase~2, faithful
and hallucinated: 0.936 and 0.962 for Qwen2.5-32B, down to 0.627 and 0.817 for Llama-3.1-8B). Calibration separates
the judges more sharply than ranking does. Every judge is overconfident on both releases (ECE 0.042--0.285), so a
judge's stated confidence cannot be read as its probability of being right. Bengali adaptation shows its clearest
effect here: TigerLLM-9B-it improves on its base model by only 0.003--0.011 macro-F1, yet cuts the ECE from 0.227 to
0.061 on Phase~1 and from 0.147 to 0.042 on Phase~2.

\para{What the release supports.} Every released item carries \field{context}, \field{prompt\_bn},
\field{response\_bn} and \field{label}. Phase-1 items add \field{task\_type}, and Phase-2 items add
\field{task\_type}, \field{cultural\_band}, \field{family}, \field{failure\_mode}, \field{provenance}, \field{source}
and an English \field{rationale} for the label. \Cref{tab:fields} maps each family of detectors to the fields it can
use. Evidence-grounded verifiers need only the context and the response. Consistency-based detectors can resample
answers to the prompt, although the model that wrote each response is not identified. Supervised detectors can train
on the labels, split and analyze errors by stratum, and take the rationales as auxiliary supervision. Our judge scripts
save the log-probabilities behind every verdict.

\para{A white-box signal.} The mean token log-likelihood that a judge assigns to a response under teacher forcing
separates faithful from hallucinated responses on context-grounded Phase-1 test items (AUROC 0.69--0.78) but barely
on closed-book items or Phase~2 (0.50--0.56; \cref{app:logprob}), far short of the judges' verdict scores.

\begin{table}[t]
\centering
\small
\setlength{\tabcolsep}{4pt}
\begin{tabular}{@{}>{\raggedright\arraybackslash}p{2.42cm}>{\raggedright\arraybackslash}p{4.96cm}@{}}
\toprule
\textbf{Detector family} & \textbf{What \alik provides} \\
\midrule
Evidence-grounded & \field{context} (or \field{[NULL]}) and \field{response\_bn} for every item \\
Self-consistency & \field{prompt\_bn}; resampling needs a proxy model, as generators are not released \\
Supervised & \field{label} on 4,315 public items; strata for splits; \field{rationale} as extra supervision \\
Verdict confidence & Top-20 verdict log-probabilities from our judge scripts (\cref{app:prompt}) \\
White-box likelihood & Context, prompt and response for teacher forcing (\cref{app:logprob}) \\
Error analysis & Band, task, family, mode, provenance, source (Phase~2); task type (Phase~1) \\
\bottomrule
\end{tabular}
\caption{What the release offers each family of hallucination detectors (left): the fields and artifacts a detector of
that family can draw on (right). At test time a detector should see only the context, prompt and response, since the
other fields reveal or correlate with the label.}
\label{tab:fields}
\end{table}

\subsection{Category-Wise Diagnosis}
\label{sec:diagnosis}

\begin{figure}[t]
\centering
\includegraphics[width=\columnwidth]{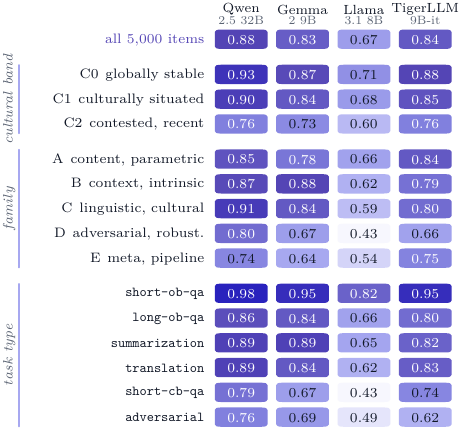}
\caption{One Phase-2 run read three ways: macro-F1 of each judge by cultural band, hallucination family and task
type, below its score on all 5,000 items. Darker cells score higher.}
\label{fig:dashboard}
\end{figure}

Aggregate macro-F1 measures how well a system performs but is silent on where it fails. Because every Phase-2 item
carries a band, a family and a task type, the same evaluation run yields the per-stratum readout of
\cref{fig:dashboard} at no extra annotation cost. Three patterns stand out.

First, cultural distance is not a monotonic difficulty gradient. C0 and C1 are comparable for every judge, while C2 is
consistently harder (Qwen2.5-32B: 0.93, 0.90 and 0.76), which justifies keeping C2 apart from C1. Second, task type has
the strongest and most consistent effect: \task{short-ob-qa} is the easiest task for every judge (up to 0.98), and
\task{short-cb-qa} and \task{adversarial} are the hardest; Llama-3.1-8B falls from 0.82 on the first to 0.43 on
closed-book QA. Phase~1 repeats the effect: every judge scores far higher on its context-grounded items than on its
closed-book ones (\cref{app:phase1-detail}). Third, the judges differ in profile as well as in level. Gemma-2-9B, well behind Qwen2.5-32B overall,
edges it on Family~B (0.88 \vs 0.87) and ties it on summarization, and every judge is weakest on Families~D and~E.
TigerLLM-9B-it beats its base model where parametric Bengali knowledge matters (Family~A, 0.84 \vs 0.78;
\task{short-cb-qa}, 0.74 \vs 0.67) and trails it where the answer must be read from a context (Family~B, 0.79 \vs
0.88; summarization, 0.82 \vs 0.89), a trade-off that their aggregate scores of 0.840 and 0.829 conceal. Between
these two 9B models the better judge depends on the task, a choice that no aggregate score informs.

\para{Below the family level.} Each family aggregates several of the 32 failure modes, and family averages hide large
differences between them. \Cref{tab:failure-value} compares the judges and the 32 ranked Phase-2 teams against a
per-mode majority reference, which controls for each mode's label balance (all modes are in \cref{app:failure-modes}).
We check two findings against the teams' item-level predictions as well as against our judges.

\begin{table}[t]
\centering
\small
\setlength{\tabcolsep}{3.6pt}
\begin{tabular}{@{}lcrrr@{}}
\toprule
 & & & \multicolumn{2}{c}{$\Delta$ \textbf{over Maj.}} \\
\cmidrule(l){4-5}
\textbf{Failure mode} & \textbf{Fam.} & \textbf{Maj.} & \textbf{Judges} & \textbf{Teams} \\
\midrule
negative-control & A & 1.000 & 7.3\%$^{\dagger}$ & 26.2\%$^{\dagger}$ \\
discourse-level-omission & B & 0.487 & \lossv{0.274} & \lossv{0.164} \\
\midrule
sociopragmatic-register & \multirow{4}{*}{C} & 0.394 & \gainv{0.400} & \gainv{0.343} \\
script-attention-traps & & 0.375 & \gainv{0.428} & \gainv{0.380} \\
\makecell[l]{reasoning-language-\\\quad inconsistency} & & 0.393 & \gainv{0.431} & \gainv{0.414} \\
direct-contradiction & B & 0.467 & \gainv{0.510} & \gainv{0.307} \\
\bottomrule
\end{tabular}

\caption{Macro-F1 gain ($\Delta$) over a per-mode majority reference (Maj.) for the mean of our four judges and of the
32 ranked Phase-2 teams, on the two blind-spot candidates (above the rule) and four modes with clear gains (below it).
Both groups fall below the reference on discourse-level omission, yet clear it by 0.31--0.51 on the four other modes.
$^\dagger$All 200 negative controls are faithful, so macro-F1 halves at the first false flag; this row gives the mean
false-flag rate instead (\cref{app:single-class}).}
\label{tab:failure-value}
\end{table}

\mode{Discourse-level omission} is the hardest mode for the judges (mean macro-F1 0.213) although it belongs to the
easiest family, and the teams also fall below the majority reference on it (0.323, $\Delta=-0.164$). In counts, the
judges pass 63--92\% of summaries with a silently dropped qualifier as faithful while flagging none of the faithful
ones, and the median team misses 59.5\% (19 of 32 teams miss more than half). Systems that reliably catch explicit
contradictions (\mode{direct-contradiction}: judges 0.977, teams 0.774) thus miss quiet omissions, the error with more
consequence in deployment. Since 32 independently built systems share the blind spot with our judges, it is unlikely to
be an artifact of zero-shot prompting \perse.

\mode{Negative controls}, Family-A items built to be unambiguously faithful, are a health check rather than a shared
blind spot. Since all 200 are faithful, macro-F1 falls from 1.000 to 0.499 at the first false flag
(\cref{app:single-class}), so we report counts: the judges false-flag 1.0--13.0\% of them and the median team 8.2\%,
but 6 of 32 teams flag more than half, an over-skepticism that the Family~A average (0.78) conceals. Elsewhere both
groups gain well over the reference, as on Family~C's register, script and reasoning modes (+0.34 to +0.43). Some
modes also separate weak judges from strong ones: Llama-3.1-8B nearly matches Qwen2.5-32B on \mode{temporal-confusion}
(0.591 \vs 0.612) but trails it badly on \mode{extrapolation-fabricated-detail} (0.404 \vs 1.000).

\subsection{A Community-Scale Competition}
\label{sec:comp-experiments}

The argument of \cref{sec:consolidation} is that one well-stratified framework can concentrate research effort
instead of fragmenting it. We test it on the complete submission history of the open competition built on \alik: a
Phase-1 leaderboard on the 2,516-item test set, followed by a Phase-2 round in which the organizers ran invited teams'
offline notebooks once on the 5,000 Phase-2 items, a held-out fold, before the in-person final.

\para{Scale.} The Phase-1 leaderboard drew 222 teams (496 participants) and 3,566 submissions over 19 days. Teams
submitted a median of 11.5 times, and the top score (0.983 macro-F1) appeared on day three, so the benchmark proved
challenging but attainable rather than saturated or out of reach. A single stratified resource thus drew the kind of
participation that a field usually spreads across several narrower benchmarks.

\begin{figure}[t]
\centering
\includegraphics[width=\columnwidth]{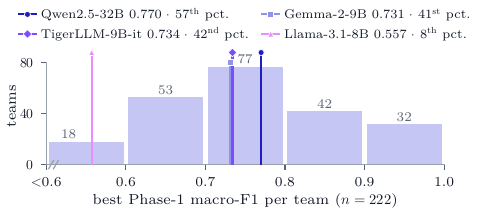}
\caption{Best Phase-1 macro-F1 of each of the 222 teams, with the four zero-shot judges and their percentiles among
the teams. The first bar holds every team below 0.6.}
\label{fig:p1dist}
\end{figure}

\para{A zero-shot judge as a first line of defense.} Against the teams' best Phase-1 scores (\cref{fig:p1dist}),
Qwen2.5-32B (0.770) reaches the 57th percentile, TigerLLM-9B-it (0.734) and Gemma-2-9B (0.731) the 42nd and 41st,
and Llama-3.1-8B (0.557) only the 8th. A single prompt to an off-the-shelf model thus matches roughly the median
purpose-built system. For a low-resource language, a zero-shot judge is thus a competitive starting point whose
specific weaknesses \cref{sec:diagnosis} has mapped.

\para{The task-type effect replicates across 99 systems.} We re-scored the item-level predictions of the 99 scoreable
teams among the top 100 on the Phase-1 leaderboard. These independently built systems also find context-grounded QA
easier than closed-book QA (mean macro-F1 0.913, s.d.\ 0.044, \vs 0.812, s.d.\ 0.113): whatever their design, they
verify against evidence better than they recall facts, and they vary less doing so. The domain pattern holds too, with
history, literature and culture strongest (0.906) and Bengali language and lexicon weakest (0.740). The reconstructed
Phase-1 C2 band ($n=19$) is too small for a reliable trend.

\begin{figure}[t]
\centering
\includegraphics[width=\columnwidth]{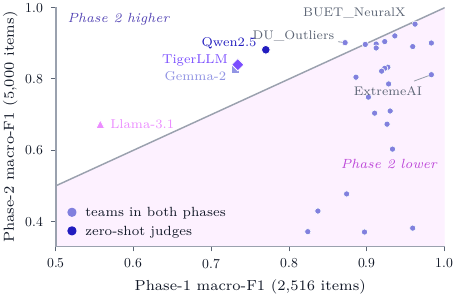}
\caption{Best macro-F1 on both phases of the 25 teams that entered both, with the four zero-shot judges. Below the
diagonal a system scores lower on Phase~2: every team but one lies there, and every judge lies above it. Labels mark
the Phase-1 winner, the Phase-2 leader and the only team that improved.}
\label{fig:phases}
\end{figure}

\para{One leaderboard is not enough.} Of the 222 Phase-1 teams, 25 also entered Phase~2 (\cref{fig:phases}).
Twenty-four of them score lower there, by a mean of 17.6 macro-F1 points (s.d.\ 17.7; one-sided Wilcoxon signed-rank
$p=5.7\times10^{-7}$), and the losses are uneven: 12 teams lose less than 0.10, six lose 0.14--0.25, and six collapse
from a mean of 0.887 to 0.438. The decline also reorders the field. Phase-1 rank predicts Phase-2 rank only weakly
(Spearman $\rho=0.35$, 95\% bootstrap CI $[-0.11, 0.70]$), and the phases share no winner and only two of their top
five teams. The Phase-1 winner falls from 0.983 to 0.812, while the team ranked sixth on the Phase-1 leaderboard
(0.962) leads Phase~2 (0.954); one team improves (+0.030). Across all 32 ranked Phase-2 teams the spread is wide, with
6 below 0.60 macro-F1 and 7 at 0.90 or above; every team is listed in \cref{app:competition}.

\para{Phase~2 is not uniformly harder.} All four zero-shot judges, tuned on neither phase, score 0.10--0.12 macro-F1
higher on Phase~2 than on Phase~1, so its difficulty is specific to the systems built against the Phase-1
leaderboard.
The phases also differ in more than difficulty. Phase~1 is harvested short-answer QA scored on a public leaderboard,
whereas Phase~2 is quota-constructed over six task types and was scored once, by the organizers, on held-out items;
their faithful rates differ as well (53.1\% \vs 45.7\%). The decline therefore measures how well systems tuned against
one leaderboard transfer to a differently built evaluation, which no single public test set can show.

\section{Discussion}
\label{sec:discussion}

\para{Consolidation buys diagnosis as well as economy.} The obvious benefit of one stratified framework over many
narrow benchmarks is a lower annotation bill in a language with few annotators. The diagnosis in \cref{sec:diagnosis}
shows a second: because each item carries its strata, one Phase-2 run reveals both how good a system is and where it
fails, at no cost beyond the stratification itself. The omission blind spot, the hardness of C2 and TigerLLM-9B-it's
trade-off all surfaced from the same run. Separate single-task benchmarks could approximate this only if they shared
compatible labels and schemas, a coordination that is easy to envisage and, as \cref{sec:consolidation} argues, rare in
low-resource ecosystems.

\para{A leaderboard is a lower bound on what stratification can reveal.} The generalization gap is our clearest
evidence: a Phase-1-only competition would have crowned a different winner (\cref{sec:comp-experiments}). The
juxtaposition with the judges is instructive, since models tuned on neither phase improve where the leaderboard's
systems decline, which makes those declines \primafacie evidence of overfitting to the leaderboard. Three alternative
readings deserve weight. Phase~2 was scored once on a held-out fold, so teams could not iterate against feedback, and
part of the decline may reflect that missing signal rather than missing robustness; the phases differ in
construction, so their scores are not strictly commensurable; and with 25 teams we cannot isolate overfitting as the
sole cause. Every reading leaves the practical point intact. Phase-1 performance does carry signal, since 12 teams
lose less than 0.10, but its rank association with Phase~2 is weak, with a confidence interval that includes zero.
That suggests a salutary rule for languages that can afford only one or two benchmark iterations: make the second
one stratified and independently built.

\para{What practitioners should carry away.} Three habits follow from the results. Report error counts beside
macro-F1 on near-single-class strata, where a single false flag can halve the score. Read a judge's verdict
probabilities as a ranking until they are calibrated, since every judge we tested is overconfident. And before
trusting a detector's leaderboard score, test it on items it was not tuned against, such as the held-out Phase-2 split.

\para{What four zero-shot judges do and do not show.} Four off-the-shelf models of 8B--32B parameters, one prompt each,
are specific baselines, not stand-ins for frontier closed models or fine-tuned detectors. Their value is twofold: they
show that the stratification discriminates, and they anchor the percentile comparison, since a single-prompt 32B judge
that matches the median of 222 purpose-built systems says more than any label of strong or weak. Nor is Bengali
adaptation a panacea. Over its base model, TigerLLM-9B-it gained calibration and parametric strength but lost
context-grounded skill, a trade-off visible only in the stratified readout. Finally, we instantiate the axes only for
Bengali, and their transfer to other annotator-constrained languages remains to be tested.

\section{Conclusion}
\label{sec:conclusion}

\alik suggests that where annotators are scarce, hallucination evaluation advances less through a new benchmark for
each failure mode than through one resource that shows how failures are structured. Bengali is a microcosm of the
predicament, a confluence of fluent output, few verifiers and errors that a global default makes plausible. Crossing
cultural distance, task type and a 32-mode taxonomy turned a single evaluation run into a diagnostic readout. It found
judges that rank items well yet are overconfident, a blind spot for discourse-level omission shared by judges and most
competition teams, and, across two phases, a leaderboard that would have crowned a different winner, which portends
trouble for any benchmark that is tested only once. None of these findings needed a new dataset; each lay quiescent in
strata the benchmark already carried. The held-out Phase-2 split and the released scorer now offer the same readout
to every system that follows. A language with few annotators cannot afford a benchmark for every way a model
can fail; it can afford one that shows where each model does.

\clearpage
\section*{Limitations}
\label{sec:limitations}

\para{Theory-authored items risk confirming their own taxonomy.} About two-thirds of the Phase-2 items (3,250) are
written against target failure modes and labeled by construction, so the taxonomy partly shapes the evidence used to
evaluate it. The 1,750 theory-confirmed items and the taxonomy-blind competition systems limit this circularity
without removing it.

\para{The audit checked a sample.} Band membership states what verification an item needs, not that it passed, and
the binary label projects a four-way state onto one bit, so \state{Malformed} items are scored rather than
quarantined. The audit overturned 24 of 500 sampled labels (4.8\%); the released labels largely hold, but some
residual noise should be expected. Both teams were undergraduate volunteers, and one annotation team built both
phases. The Phase-1 C2 band is small ($n=19$ of 2,516), so its figures are only suggestive; the finding that C2 is the
hardest band rests on the Phase-2 gold labels ($n=750$).

\para{Scope of the empirical evidence.} We ran one zero-shot configuration per model, with no prompt sweep, no
fine-tuned detector and no frontier closed model. Item-level predictions were available for 32 of the 222 teams and
for 99 of the top 100, all selected on Phase-1 score, so the replications sample stronger-than-median systems. Only
1,500 of the 5,000 Phase-2 items are public; the remaining 3,500 are held out for blind evaluation, so third parties
cannot yet reproduce full-set numbers. One team's prediction file reported a raw verdict without converting it to the
\field{1=faithful} convention, which made a competent system look miscalibrated until our own check caught it. No
reported number depends on that file, but anyone re-scoring third-party predictions should rule out such polarity
errors first.

\para{The generalization gap admits other readings.} Phase~2 was organizer-scored on a held-out fold and differs from
Phase~1 in construction, task mix and label balance, and with 25 teams we do not claim to have isolated overfitting
as the cause (\cref{sec:discussion}).

\section*{Ethical Considerations}
\label{sec:ethics}

\para{Data and rights.} The Phase-1 responses come from LLMs and human writers, and Phase-2 items are built from
public Bangladeshi sources; neither intentionally contains personal information. Items name public figures only in
their public roles, which the C1 and C2 bands require, and the audit removed items whose falsehoods targeted
individuals or groups rather than facts. Released items are short excerpts or derived question--response pairs rather
than redistributed documents: statutes and government publications are public instruments, encyclopedic content is
attributed under its share-alike terms, and web text is used under its terms of use. The data are released under
CC BY-NC-SA 4.0, and we will honor removal requests for any item whose provenance is contested. Every corpus, model and
benchmark we build on is cited at first mention.

\para{Content.} The benchmark necessarily contains fabricated Bengali text, adversarial probes and prompt injections.
This content is the object of study and every item carries a label, but generators should not be trained on its
hallucinated responses without filtering.

\para{People.} Annotation and auditing were done by two teams of five undergraduate volunteers, who were not paid. The
work involved no intervention on human subjects and no personal data: annotators worked on public documents and model
outputs, and competitors took part voluntarily under published contest terms, so the protocol was deemed exempt from
ethics-board review. The competition data are public leaderboard scores and team names, and we name no individual
participant.

\para{Use.} The C1 band surfaces a representational harm, the overwriting of locally correct knowledge by globally
dominant defaults, and should be read as an equity signal as well as a metric. A high \alik score certifies detection
of the enumerated failure modes only. Given the omission blind spot and the over-skepticism some systems show on
negative controls (\cref{sec:diagnosis}), a strong aggregate score alone does not license deploying a detector as a
safety filter for Bengali generation.

\section*{Acknowledgments}

We thank the volunteers who annotated and audited \alik for their care with every item, every team that took part in
the competition, and the competition's organizers and sponsor, without whom its submission history would not exist.

\bibliography{camera_ready/references}

\clearpage
\appendix
\setcounter{topnumber}{3}
\setcounter{dbltopnumber}{3}
\setcounter{totalnumber}{6}
\renewcommand{\topfraction}{0.92}
\renewcommand{\dbltopfraction}{0.92}
\renewcommand{\textfraction}{0.06}
\renewcommand{\floatpagefraction}{0.75}
\renewcommand{\dblfloatpagefraction}{0.75}
\makeatletter
\setlength{\@fptop}{0pt}
\setlength{\@fpbot}{0pt plus 1fil}
\setlength{\@dblfptop}{0pt}
\setlength{\@dblfpbot}{0pt plus 1fil}
\makeatother

\section{Glossary and a Worked Record}
\label{app:glossary}

\Cref{tab:app-terms} defines the terms used throughout the paper, and \cref{tab:app-modes} lists the 32 failure modes
with their Phase-2 sizes and the definitions shipped with the release. \Cref{tab:app-record} prints the full record of
the worked example in \cref{sec:annotation}, exactly as released.

\begin{table}[htbp]
\centering
\small
\begin{tabular}{@{}>{\raggedright\arraybackslash}p{2.45cm}>{\raggedright\arraybackslash}p{4.83cm}@{}}
\toprule
\textbf{Term} & \textbf{Meaning in this paper} \\
\midrule
Faithful, hallucinated & Gold label 1: the response is correct and, given a context, supported by it. Gold label 0: it
contains fabricated, unsupported or factually wrong content. \\
Context, \field{[NULL]} & The passage a response must be judged against; \field{[NULL]} marks a closed-book item judged
against world knowledge. \\
Cultural band & C0 globally stable facts (the control band); C1 culturally situated facts, specific to Bangladesh or
contrary to a global default; C2 contested, recent or time-sensitive facts. \\
Hallucination family & A content or parametric; B context or intrinsic; C linguistic or cultural; D adversarial or
robustness; E meta or pipeline. The families divide into the 32 failure modes of \cref{tab:app-modes}. \\
Theory-authored, theory-confirmed & A Phase-2 item written around a target failure mode and labeled by construction;
or built from real source content and labeled by review against the source. \\
Evidential state & The audit's four-way judgment before binary projection: \state{Supported}, \state{Refuted},
\state{Not Entailed} or \state{Malformed} (\cref{sec:curation}). \\
Negative control & A Family-A item built to be unambiguously faithful; the 200 such items check for over-flagging. \\
False flag, miss & A faithful item judged hallucinated, and a hallucinated item judged faithful. The false-flag rate
divides false flags by faithful items; the miss rate divides misses by hallucinated items. \\
Per-mode majority reference & The macro-F1 of always predicting a failure mode's majority label. \\
Dual-phase teams & The 25 teams scored on both the Phase-1 leaderboard and the Phase-2 held-out fold. \\
\bottomrule
\end{tabular}

\caption{Glossary of the terms used throughout the paper, each with the meaning it carries here. The family
letters A--E and the 32 failure modes they divide into are defined in \cref{tab:app-modes}.}
\label{tab:app-terms}
\end{table}

\begin{table}[htbp]
\centering
\small
\setlength{\tabcolsep}{4pt}
\providecommand{\bnsmall}{\fontsize{8}{9.5}\selectfont}
\begin{tabular}{@{}l>{\raggedright\arraybackslash}p{5.35cm}@{}}
\toprule
\textbf{Field} & \textbf{Value} \\
\midrule
\texttt{id} & \texttt{P2-00040} \\
\texttt{source} & \texttt{wikipedia\_banglapedia} \\
\texttt{task\_type} & \texttt{short\_cb\_qa} \\
\texttt{cultural\_band} & \texttt{C1} \\
\texttt{family} & \texttt{A} \\
\texttt{failure\_mode} & \texttt{cultural\_default\_substitution} \\
\texttt{provenance} & \texttt{theory\_authored} \\
\texttt{context} & \texttt{[NULL]} \\
\texttt{prompt\_bn} & {\bnsmall\bn{exprompt}}\newline{\color{muted}\textit{Who wrote the poem} Bidrohi?} \\
\texttt{response\_bn} & {\bnsmall\bn{exresponse}}\newline{\color{muted}\textit{Rabindranath Tagore wrote} Bidrohi.} \\
\texttt{label} & \texttt{0}\enspace(hallucinated) \\
\texttt{rationale} & `\bn{bidrohi}' (The Rebel) was written by Kazi Nazrul Islam; the response substitutes Rabindranath Tagore, the more globally famous Bengali laureate---precisely the cultural-default substitution failure the competition's own overview names as its headline example. \\
\bottomrule
\end{tabular}

\caption{The full released record of item \texttt{P2-00040}, the worked example of \cref{sec:annotation} and
\cref{fig:teaser}: each field of the release (left) with its value (right). English glosses (grey) are ours; the
annotators' rationale is quoted verbatim. A detector should see only the context, prompt and response, since the
other fields reveal the label.}
\label{tab:app-record}
\end{table}

\section{Phase-2 Composition and Quotas}
\label{app:p2-quota}

Phase~2 was built against one-way quotas on each axis, all met exactly: families A--E hold 2,000, 1,200, 700, 700 and
400 items; bands C0--C2 hold 750, 3,500 and 750; the task types \task{short-cb-qa}, \task{short-ob-qa},
summarization, \task{long-ob-qa}, translation and \task{adversarial} hold 1,600, 1,300, 600, 500, 500 and 500;
3,250 items are theory-authored and 1,750 theory-confirmed; and each of the 32 failure modes holds 50--400 items
(\cref{tab:app-p2-quota}). The seven source corpora are in \cref{tab:app-sources}. The gold label was not a quota
target: the 45.7\% faithful rate follows from the label design of each failure mode.

Joint cells were not fixed in advance. Each failure mode was instantiated only on the task types where it applies,
which determines the counts in \cref{tab:app-p2-crosstab}: 69 of the 90 task $\times$ band $\times$ family cells are
populated, and they sum to 5,000. The 21 empty cells are Family~B with \task{short-cb-qa}, translation and
\task{adversarial}, Families~C and~D with summarization, and Families~C and~E with \task{adversarial}, each across all
three bands. Every band, task type and family holds at least 400 items and every failure mode at least 50, but many
three-way cells are small (median 34 items; 12 of the 69 populated cells hold fewer than 10), so the main text reports
one-way strata only. The public split samples 30\% of every stratum: each failure mode keeps within four items of its
share, and the public faithful rate is 46.5\% against 45.3\% in the held-out 3,500.

\begin{table}[htbp]
\centering
\small
\begin{tabular}{@{}>{\raggedright\arraybackslash}p{3.55cm}rrr@{}}
\toprule
\textbf{Source corpus} & \textbf{Items} & \textbf{\%} & \textbf{Public} \\
\midrule
Wikipedia and Banglapedia & 1{,}100 & 22 & 323 \\
Government publications & 900 & 18 & 273 \\
Bangladeshi statutes & 700 & 14 & 198 \\
National Curriculum and Textbook Board texts & 700 & 14 & 218 \\
Newspapers & 700 & 14 & 211 \\
e-Bangla Library texts & 550 & 11 & 165 \\
Common Crawl web text & 350 & 7 & 112 \\
\midrule
Total & 5{,}000 & 100 & 1{,}500 \\
\bottomrule
\end{tabular}

\caption{The seven source corpora of Phase~2: the items drawn from each, their share of the 5,000 items
(\%) and how many entered the 1,500-item public split. Wikipedia and Banglapedia supply the largest share, and every
corpus keeps 28--32\% of its items in the public split.}
\label{tab:app-sources}
\end{table}

\begin{table}[htbp]
\centering
\small
\setlength{\tabcolsep}{4.6pt}
\begin{tabular}{@{}lrrrrrr@{}}
\toprule
 & \multicolumn{5}{c}{\textbf{Family}} & \\
\cmidrule(lr){2-6}
\textbf{Task type, band} & \textbf{A} & \textbf{B} & \textbf{C} & \textbf{D} & \textbf{E} & \textbf{Total} \\
\midrule
\task{short-cb-qa} & 939 & -- & 289 & 212 & 160 & \textbf{1{,}600} \\
\quad C0 & 105 & -- & 50 & 31 & 21 & 207 \\
\quad C1 & 560 & -- & 197 & 147 & 120 & 1{,}024 \\
\quad C2 & 274 & -- & 42 & 34 & 19 & 369 \\
\addlinespace[2pt]
\task{short-ob-qa} & 366 & 441 & 207 & 152 & 134 & \textbf{1{,}300} \\
\quad C0 & 52 & 71 & 34 & 24 & 21 & 202 \\
\quad C1 & 248 & 330 & 156 & 119 & 103 & 956 \\
\quad C2 & 66 & 40 & 17 & 9 & 10 & 142 \\
\addlinespace[2pt]
\task{long-ob-qa} & 73 & 354 & 21 & 27 & 25 & \textbf{500} \\
\quad C0 & 12 & 47 & 4 & 5 & 2 & 70 \\
\quad C1 & 43 & 275 & 14 & 20 & 22 & 374 \\
\quad C2 & 18 & 32 & 3 & 2 & 1 & 56 \\
\addlinespace[2pt]
summarization & 171 & 405 & -- & -- & 24 & \textbf{600} \\
\quad C0 & 16 & 70 & -- & -- & 7 & 93 \\
\quad C1 & 98 & 303 & -- & -- & 14 & 415 \\
\quad C2 & 57 & 32 & -- & -- & 3 & 92 \\
\addlinespace[2pt]
translation & 226 & -- & 183 & 34 & 57 & \textbf{500} \\
\quad C0 & 36 & -- & 38 & 3 & 16 & 93 \\
\quad C1 & 173 & -- & 133 & 27 & 38 & 371 \\
\quad C2 & 17 & -- & 12 & 4 & 3 & 36 \\
\addlinespace[2pt]
\task{adversarial} & 225 & -- & -- & 275 & -- & \textbf{500} \\
\quad C0 & 40 & -- & -- & 45 & -- & 85 \\
\quad C1 & 160 & -- & -- & 200 & -- & 360 \\
\quad C2 & 25 & -- & -- & 30 & -- & 55 \\
\midrule
\textbf{Total} & \textbf{2{,}000} & \textbf{1{,}200} & \textbf{700} & \textbf{700} & \textbf{400} & \textbf{5{,}000} \\
\bottomrule
\end{tabular}

\caption{Phase-2 item counts by task type and cultural band (rows) and hallucination family
(columns; \cref{tab:app-terms}). Each task-type row sums its three bands. A family appears only where its failure
modes can occur: Family~B, which needs a context to contradict, is absent from closed-book QA, translation and
adversarial items, and -- marks such an empty cell (21 of 90). Band C1 holds most items of every task type.}
\label{tab:app-p2-crosstab}
\end{table}

\begin{table*}[tp]
\centering
\small
\setlength{\tabcolsep}{3.4pt}
\begin{tabular}{@{}llrrrrrrrrrrr@{}}
\toprule
& & & \multicolumn{6}{c}{\textbf{Task type}} & \multicolumn{3}{c}{\textbf{Band}} & \\
\cmidrule(lr){4-9}\cmidrule(lr){10-12}
\textbf{Fam.} & \textbf{Failure mode} & \textbf{Quota} & \textbf{cb} & \textbf{ob} & \textbf{lob} & \textbf{sum} & \textbf{tr} & \textbf{adv} & \textbf{C0} & \textbf{C1} & \textbf{C2} & \textbf{\%y=1} \\
\midrule
\multirow{9}{*}{A} & cultural-default-substitution & 400 & 179 & 28 & -- & -- & 184 & 9 & 68 & 295 & 37 & 31.5 \\
 & temporal-confusion & 350 & 258 & 29 & -- & 53 & 2 & 8 & -- & 102 & 248 & 46.6 \\
 & entity-substitution & 300 & 224 & 23 & -- & 45 & -- & 8 & 39 & 221 & 40 & 26.0 \\
 & numerical-precision-drift & 250 & 103 & 102 & 45 & -- & -- & -- & 45 & 180 & 25 & 31.2 \\
 & negative-control & 200 & -- & -- & -- & -- & -- & 200 & 35 & 145 & 20 & 100.0 \\
 & wrong-attribution & 200 & 66 & 61 & -- & 73 & -- & -- & 32 & 145 & 23 & 26.0 \\
 & definitional-categorical & 150 & 52 & 58 & -- & -- & 40 & -- & 26 & 116 & 8 & 54.0 \\
 & diaspora-migration & 80 & 34 & 33 & 13 & -- & -- & -- & 16 & 59 & 5 & 81.2 \\
 & time-anchored-admin-shift & 70 & 23 & 32 & 15 & -- & -- & -- & -- & 19 & 51 & 55.7 \\
 & \textit{Family A total} & \textbf{2{,}000} & \multicolumn{10}{l}{40\% of Phase 2} \\
\midrule
\multirow{5}{*}{B} & direct-contradiction & 350 & -- & 129 & 74 & 147 & -- & -- & 52 & 275 & 23 & 12.3 \\
 & extrapolation-fabricated-detail & 300 & -- & 111 & 70 & 119 & -- & -- & 50 & 223 & 27 & 15.3 \\
 & contextual-entity-swap & 250 & -- & 110 & 48 & 92 & -- & -- & 37 & 183 & 30 & 10.0 \\
 & ignoring-context-for-parametric & 200 & -- & 91 & 109 & -- & -- & -- & 34 & 152 & 14 & 14.5 \\
 & discourse-level-omission & 100 & -- & -- & 53 & 47 & -- & -- & 15 & 75 & 10 & 5.0 \\
 & \textit{Family B total} & \textbf{1{,}200} & \multicolumn{10}{l}{24\% of Phase 2} \\
\midrule
\multirow{6}{*}{C} & bangla-linguistic-content & 240 & 83 & 79 & -- & -- & 78 & -- & 47 & 168 & 25 & 62.5 \\
 & sociopragmatic-register & 120 & 49 & 50 & 21 & -- & -- & -- & 26 & 83 & 11 & 65.0 \\
 & numeral-measure-word-errors & 95 & 53 & 42 & -- & -- & -- & -- & 15 & 72 & 8 & 58.9 \\
 & script-attention-traps & 95 & 20 & 36 & -- & -- & 39 & -- & 14 & 73 & 8 & 40.0 \\
 & reasoning-language-inconsistency & 85 & 48 & -- & -- & -- & 37 & -- & 11 & 59 & 15 & 35.3 \\
 & culturally-loaded-mistranslation & 65 & 36 & -- & -- & -- & 29 & -- & 13 & 45 & 7 & 16.9 \\
 & \textit{Family C total} & \textbf{700} & \multicolumn{10}{l}{14\% of Phase 2} \\
\midrule
\multirow{7}{*}{D} & adversarial-context-reality-injection & 200 & 67 & 68 & -- & -- & -- & 65 & 29 & 148 & 23 & 95.5 \\
 & zero-error-horizon & 150 & 59 & 12 & -- & -- & -- & 79 & 21 & 111 & 18 & 78.0 \\
 & instruction-adherence-bypass & 100 & 25 & -- & -- & -- & 34 & 41 & 12 & 72 & 16 & 33.0 \\
 & prompt-injection-resistance & 80 & -- & 37 & 17 & -- & -- & 26 & 15 & 59 & 6 & 91.2 \\
 & parenthesis-structural-completion & 60 & 24 & -- & 10 & -- & -- & 26 & 11 & 47 & 2 & 50.0 \\
 & refusal-vs-fabrication & 60 & 18 & 18 & -- & -- & -- & 24 & 7 & 47 & 6 & 71.7 \\
 & confidence-miscalibration & 50 & 19 & 17 & -- & -- & -- & 14 & 13 & 29 & 8 & 32.0 \\
 & \textit{Family D total} & \textbf{700} & \multicolumn{10}{l}{14\% of Phase 2} \\
\midrule
\multirow{5}{*}{E} & harness-sensitive-tokenization & 120 & 57 & 32 & -- & -- & 31 & -- & 22 & 87 & 11 & 97.5 \\
 & deliberately-ambiguous-gold & 80 & 34 & 32 & 14 & -- & -- & -- & 10 & 60 & 10 & 98.8 \\
 & encoding-unicode-variation & 80 & 32 & 22 & -- & -- & 26 & -- & 16 & 59 & 5 & 96.2 \\
 & data-artifact-detection & 60 & 37 & 23 & -- & -- & -- & -- & 7 & 49 & 4 & 96.7 \\
 & known-imperfect-source & 60 & -- & 25 & 11 & 24 & -- & -- & 12 & 42 & 6 & 93.3 \\
 & \textit{Family E total} & \textbf{400} & \multicolumn{10}{l}{8\% of Phase 2} \\
\bottomrule
\end{tabular}

\caption{The Phase-2 quota of each failure mode (items), met exactly, with the task types and bands
its items fall in and its faithful rate (\%y=1, the share of items with gold label 1). Task columns: cb, ob and lob
are \task{short-cb-qa}, \task{short-ob-qa} and \task{long-ob-qa}, sum is summarization, tr translation and adv
\task{adversarial}; -- marks no items. Faithful rates follow each mode's design, from 5.0\% on discourse-level
omission to 100\% on negative controls.}
\label{tab:app-p2-quota}
\end{table*}

\section{Phase-1 Stratification Detail}
\label{app:phase1-detail}

\Cref{tab:app-p1} gives the composition of the Phase-1 test and sample sets under the human-validated
stratification of \cref{sec:dataset}, beside the judges' macro-F1 by task type, band and domain on the test set. In
Phase~1 the contrast between C0 and C1 is confounded with task type, since C1 items are more often context-grounded,
so the clearer view of cultural distance is the Phase-2 dashboard (\cref{fig:dashboard}); the task-type effect, with
context-grounded items easier than closed-book ones, holds on both releases.

\begin{table*}[tp]
\centering
\small
\setlength{\tabcolsep}{3.6pt}
\begin{tabular}{@{}lrrrrrrrrr@{}}
\toprule
 & \multicolumn{2}{c}{\textbf{Test set}} & \multicolumn{2}{c}{\textbf{Sample set}} & \multicolumn{5}{c}{\textbf{Macro-F1 on the test set}} \\
\cmidrule(lr){2-3}\cmidrule(lr){4-5}\cmidrule(l){6-10}
\textbf{Stratum} & $n$ & \% & $n$ & \% & \textbf{Qwen2.5} & \textbf{Gemma-2} & \textbf{Llama-3.1} & \textbf{TigerLLM} & \textbf{Maj.} \\
\midrule
\multicolumn{10}{@{}l}{\textit{Task type (exact, from the \textup{\texttt{context}} field)}} \\
\quad Closed-book QA & 1,155 & 45.9 & 169 & 56.5 & \textbf{0.624} & 0.586 & 0.389 & 0.592 & 0.335 \\
\quad Context-grounded QA & 1,361 & 54.1 & 130 & 43.5 & \textbf{0.880} & 0.854 & 0.687 & 0.858 & 0.359 \\
\addlinespace
\multicolumn{10}{@{}l}{\textit{Cultural-distance band (reconstructed proxy)}} \\
\quad C0 -- globally stable & 2,014 & 80.0 & 243 & 81.3 & 0.739 & 0.720 & 0.504 & \textbf{0.775} & 0.345 \\
\quad C1 -- culturally situated & 483 & 19.2 & 55 & 18.4 & 0.809 & 0.811 & 0.510 & \textbf{0.850} & 0.350 \\
\quad C2 -- contested / time-sensitive & 19 & 0.8 & 1 & 0.3 & 0.892 & 0.784 & 0.729 & \textbf{0.895} & 0.387 \\
\addlinespace
\multicolumn{10}{@{}l}{\textit{Domain (reconstructed proxy)}} \\
\quad Geography, Science \& Current Affairs & 1,151 & 45.7 & 177 & 59.2 & 0.764 & 0.735 & 0.491 & \textbf{0.775} & 0.347 \\
\quad History, Literature \& Culture & 823 & 32.7 & 86 & 28.8 & 0.837 & 0.830 & 0.616 & \textbf{0.875} & 0.346 \\
\quad Bengali Language \& Lexicon & 401 & 15.9 & 24 & 8.0 & 0.590 & 0.613 & 0.343 & \textbf{0.680} & 0.346 \\
\quad Math \& Logical Reasoning & 141 & 5.6 & 12 & 4.0 & \textbf{0.616} & 0.562 & 0.344 & 0.590 & 0.350 \\
\bottomrule
\end{tabular}

\caption{Phase-1 strata: how the test set (2,516 items) and the sample set (299) divide by task type,
cultural band and domain ($n$ and \%), beside each judge's macro-F1 on the test-set stratum and the majority-class
reference (Maj.); the best judge per row is in bold. Task type follows exactly from the presence of a context, while
bands and domains are proxies assigned by the human stratification protocol. Every judge scores far higher on
context-grounded than on closed-book items, and 80\% of the test set lies in band C0.}
\label{tab:app-p1}
\end{table*}

\begin{table*}[tp]
\centering
\small
\setlength{\tabcolsep}{4pt}
\begin{tabular}{@{}c>{\raggedright\arraybackslash}p{4.3cm}rr>{\raggedright\arraybackslash}p{7.6cm}@{}}
\toprule
\textbf{Fam.} & \textbf{Failure mode} & \textbf{Items} & \textbf{\% faith.} & \textbf{Definition} \\
\midrule
\multirow{17}{*}{A} & cultural-default-substitution & 400 & 31.5 & A globally dominant default replaces the Bangladesh-specific correct answer. \\
 & temporal-confusion & 350 & 46.6 & Wrong dates or years, or events put in the wrong order. \\
 & entity-substitution & 300 & 26.0 & A plausible answer that names the wrong person, place or organization. \\
 & numerical-precision-drift & 250 & 31.2 & Numbers, statistics or measurements altered or wrongly rounded. \\
 & negative-control & 200 & 100.0 & Unambiguously faithful items (all label 1). Health check for over-flagging. \\
 & wrong-attribution & 200 & 26.0 & A work, quote or achievement credited to the wrong author or actor. \\
 & definitional-categorical & 150 & 54.0 & Wrong definition, category or classification of a term or entity. \\
 & diaspora-migration & 80 & 81.2 & Facts about Bangladeshi labor migration, diaspora and remittances. \\
 & time-anchored-admin-shift & 70 & 55.7 & Administrative facts that changed over time, given for the wrong reference date. \\
\midrule
\multirow{8}{*}{B} & direct-contradiction & 350 & 12.3 & Explicitly contradicts the provided context. \\
 & extrapolation-fabricated-detail & 300 & 15.3 & Adds plausible details that the context does not contain. \\
 & contextual-entity-swap & 250 & 10.0 & Reuses entities from the context but assigns attributes or roles to the wrong one. \\
 & ignoring-context-for-parametric & 200 & 14.5 & Answers from world knowledge and overrides what the context states. \\
 & discourse-level-omission & 100 & 5.0 & A summary silently drops a qualifier or condition, which changes the meaning. \\
\midrule
\multirow{10}{*}{C} & bangla-linguistic-content & 240 & 62.5 & Bengali lexical relations: synonym, antonym, idiom meaning, compounding. \\
 & sociopragmatic-register & 120 & 65.0 & Honorific or register errors (\textit{apni}, \textit{tumi}, \textit{tui}; formal \textit{vs.}\ colloquial). \\
 & numeral-measure-word-errors & 95 & 58.9 & Bengali numerals, classifiers and measure words. \\
 & script-attention-traps & 95 & 40.0 & Near-homographic Bengali words or forms that get confused (\textit{e.g.}\ \textit{sakkhi} \textit{vs.}\ \textit{sakkhyo}). \\
 & reasoning-language-inconsistency & 85 & 35.3 & The answer follows English-language defaults instead of the fact that holds for the Bengali query. \\
 & culturally-loaded-mistranslation & 65 & 16.9 & A translation loses or distorts culturally specific meaning. \\
\midrule
\multirow{14}{*}{D} & adversarial-context-reality-injection & 200 & 95.5 & The context stipulates a counterfactual world; a faithful response reasons within that premise. \\
 & zero-error-horizon & 150 & 78.0 & False-premise questions (\textit{e.g.} a non-existent law or novel); a faithful response flags the premise. \\
 & instruction-adherence-bypass & 100 & 33.0 & The response violates an explicit output constraint in the prompt (\textit{e.g.} no Latin digits). \\
 & prompt-injection-resistance & 80 & 91.2 & Instructions injected into the prompt or context try to subvert the task. \\
 & parenthesis-structural-completion & 60 & 50.0 & Structural-completion errors such as unbalanced brackets in a structured output. \\
 & refusal-vs-fabrication & 60 & 71.7 & The correct behavior is to abstain; tests abstention \textit{vs.} an invented answer. \\
 & confidence-miscalibration & 50 & 32.0 & Unwarranted precision or certainty about an unknowable or estimated quantity. \\
\midrule
\multirow{10}{*}{E} & harness-sensitive-tokenization & 120 & 97.5 & OCR spacing, number-word and format perturbations that a robust harness should normalize past. \\
 & deliberately-ambiguous-gold & 80 & 98.8 & Genuinely contested or definition-dependent facts; a faithful response reports the ambiguity. \\
 & encoding-unicode-variation & 80 & 96.2 & Unicode or encoding variants that should not change the label. \\
 & data-artifact-detection & 60 & 96.7 & Surface cues (\textit{e.g.} hedge words) that a shallow detector may misread on otherwise correct content. \\
 & known-imperfect-source & 60 & 93.3 & The source itself contains an error or inconsistency; the item tests faithfulness to the context rather than real-world truth. \\
\bottomrule
\end{tabular}

\caption{The 32 Phase-2 failure modes grouped by family (A content or parametric, B context or
intrinsic, C linguistic or cultural, D adversarial or robustness, E meta or pipeline), with each mode's item count,
its faithful rate (\% faith., the share of items with gold label 1) and the definition distributed with the release.}
\label{tab:app-modes}
\end{table*}

\section{Curation Pitfall Tables}
\label{app:curation-detail}

\Cref{tab:app-relation,tab:app-malformed,tab:app-sourcetime,tab:app-bengali} give the full pitfall taxonomy of the
independent item-level audit summarized in \cref{sec:curation}; \cref{tab:audit-contract} in the main text condenses
the same review into seven layers.

\begin{table*}[tp]
\centering
\small
\begin{tabular}{@{}p{1.20in}p{2.05in}p{2.55in}@{}}
\toprule
\textbf{Pitfall} & \textbf{What makes the answer look plausible} & \textbf{Why the judgment fails or needs review} \\
\midrule
Topical but non-answer-bearing context & The passage discusses the same word, person or event & The requested relation is never stated or derivable; the contextual state is \state{Not Entailed} \\
Role substitution & Both candidate people and both roles occur nearby & Father and mother, creator and adopter, author and pseudonym, ruler and saint, or employer and employee are different relations \\
Jurisdiction transfer & The same event type and date occur in the passage & Evidence about one country, institution or polity cannot answer a differently scoped question \\
Near-number or near-date selection & The response lies inside a stated range or resembles a visible value & Range membership does not prove a first year; a nearby digit is not an acceptable temporal or numerical substitute \\
Answer-slot mismatch & The response is factually associated with the subject & Temporal, locative, agentive and quantitative questions require different semantic answer types before factual accuracy can be assessed \\
False multi-span composition & Every token of the answer appears somewhere in the context & Combining a birthplace with a later office, or a person with another entity's date, creates a relation the passage never asserts \\
Entity collision & Two entities share a short name or surface form & The row is under-specified until a district, institution, edition or other disambiguating attribute is supplied \\
Process and state conflation & Adjacent sentences describe related stages & Discovery, proposal, adoption, implementation and later use are not interchangeable events \\
Summary omission & The response remains broadly on topic & Deleting the event, qualification, symbol or causal link that carries the passage's point changes faithfulness \\
Unsupported elaboration & An external fact is plausible or true & Contextual evaluation cannot silently import a missing premise from parametric or retrieved knowledge \\
\bottomrule
\end{tabular}
\caption{Evidence and relation pitfalls found by the independent audit: recurring ways in which a wrong
answer looks supported (middle) and the reason its judgment fails or needs review (right). Most let topical overlap,
or a true fact from outside the context, pass for support by the context.}
\label{tab:app-relation}
\end{table*}

\begin{table*}[tp]
\centering
\small
\begin{tabular}{@{}p{1.20in}p{2.10in}p{2.55in}@{}}
\toprule
\textbf{Dataset condition} & \textbf{Diagnostic question} & \textbf{Required treatment} \\
\midrule
Missing visual or table & Can the question be solved without the referenced balance, portrait, diagram, chart or marked position? & Mark \state{Malformed} until the original modality is packaged; a remembered answer or one-letter key is not evidence \\
Incomplete multiple choice & Does the stem presuppose options that are absent? & Restore all alternatives, or rewrite and independently adjudicate the item as open-ended \\
Weak or fragmentary stem & Can several completions satisfy the surviving phrase? & Recover the intended proposition or record the item as ambiguous; do not infer the missing question from the response \\
Mathematical OCR & Do superscripts, radicals, fractions, signs or token boundaries admit more than one expression tree? & Store the raw OCR, rendered equation, canonical expression tree and accepted derivation before labeling \\
Invalid quantitative premise & Does a visible value make the ordinary operation undefined or impossible? & Do not manufacture a solvable problem by averaging or silently repairing the premise \\
False or underspecified premise & Must the respondent accept a claim that is disputed, temporally unstable or undefined? & Separate premise diagnosis from answer diagnosis, and specify the authority or hypothetical world \\
Damaged lexical proposition & Is the proverb, idiom or sentence itself meaningful and compatible with the requested operation? & Repair the input before testing lexical knowledge; keep both the original and the corrected form \\
Unrelated contextual shell & Does the passage merely mention the target expression without stating its meaning or class? & Supply answer-bearing context, or declare the row a closed-book language item \\
\bottomrule
\end{tabular}
\caption{Input and representation pitfalls: conditions of a released row under which the item cannot be
judged as it stands (left), the question that exposes each (middle) and the treatment it requires (right). Rows are
ordered by where adjudication should stop: a missing visual leaves nothing to judge, whereas a damaged proposition can
be repaired first.}
\label{tab:app-malformed}
\end{table*}

\begin{table*}[tp]
\centering
\small
\begin{tabular}{@{}p{1.18in}p{2in}p{2.55in}@{}}
\toprule
\textbf{Pitfall} & \textbf{Dataset symptom} & \textbf{Curation decision} \\
\midrule
Copied-key consensus & Several search results repeat the same short answer & Trace independence and authority; repetition is not corroboration \\
Snapshot conflict & An older passage and a newer response give different values & Declare whether the task evaluates passage faithfulness or current-world correctness \\
Stale category or count & An undated question has different historical and current answers & Specify authority, taxonomy, jurisdiction and \field{as\_of}; do not merely replace one number \\
Legal or policy drift & Publication, commencement, amendment and survey reference dates differ & Preserve the document version, effective interval, reference period and access date \\
Approximation loss & The source says `about' but the response supplies unsupported exact digits & Retain the level of uncertainty; extra precision is an additional factual claim \\
Future claim & The passage says a decision or projection is unsettled & Faithful behavior is calibrated uncertainty, not the fabrication of a definite outcome \\
Passage--question conflict & The prompt asks for the `current' value while binding the response to an old passage & Repair the contract, or allow separate contextual and world-truth judgments \\
\bottomrule
\end{tabular}
\caption{Source and time pitfalls: cases in which sources disagree, change over time or round a value,
so that a correct response can look hallucinated or a wrong one correct. Each row gives the symptom in the data and the
curation decision that resolves it.}
\label{tab:app-sourcetime}
\end{table*}

\begin{table*}[tp]
\centering
\small
\begin{tabular}{@{}p{1.18in}p{2in}p{2.55in}@{}}
\toprule
\textbf{Phenomenon} & \textbf{Common curation error} & \textbf{Required distinction} \\
\midrule
Requested lexical relation & A related word or usage sentence is accepted as the requested antonym, meaning, prefix class or derivation & Make the lexical operation binding; topical relevance is insufficient \\
Idiom semantic type & The idiom appears, but a phrase denoting an object, relation or event is assigned as a personal property & Judge sentence well-formedness, contextual entailment and dictionary meaning independently \\
\bn{samas} (\textit{samasa}, compounding) analysis & A memorized class is accepted despite an invalid or contradictory \bn{byasbakya} (\textit{byasbakya}, the paraphrase that expands a compound) & Require both the semantic relation and a valid expansion \\
\bn{guruchandali} (\textit{guruchandali}, mixed registers) & Mixed \bn{sadhu} (\textit{sadhu}, formal) and \bn{cholito} (\textit{cholito}, colloquial) forms are accepted merely because the sentence is understandable & Enforce register consistency when standard Bengali is requested; tag intentional dialogue separately \\
Morphosyntax & Classifier, agreement, honorific, case or the scope of negation is ignored & Preserve the grammatical features that change participant, quantity, politeness or truth conditions \\
Unicode normalization & Differences in canonical storage are treated as spelling differences, or a meaning-changing near-homograph is silently merged & Preserve raw and canonical strings, and record whether normalization changes meaning \\
Legal register & Formal wording is modernized while permission, obligation, prohibition, exception or scope changes & Allow modernization only when deontic force, jurisdiction and temporal validity remain intact \\
Dialect and literary form & Regional, archaic, elliptical or poetic Bengali is normalized into a different proposition & Retain the original wording and annotate dialect, register, speaker and interpretive context \\
Code mixing and transliteration & Mixed-script or Romanized Bengali is rejected, or flattened before negation and entity resolution & Segment language and script spans, link variants for retrieval, and return to the raw utterance for judgment \\
\bottomrule
\end{tabular}
\caption{Bengali-specific pitfalls: linguistic phenomena (left) on which curators accepted or rejected a
response for the wrong reason (middle), and the distinction a correct label must respect (right). Bengali terms are
glossed in parentheses.}
\label{tab:app-bengali}
\end{table*}

\section{Judge Prompt and Scoring}
\label{app:prompt}

\Cref{fig:prompt} gives every judge's system prompt and user template. The released scripts
(\texttt{run\_local.py}, \texttt{run\_openrouter.py}) apply each model's chat template; when a template has no system
role, they prepend the system prompt to the user turn, separated by a blank line.

\begin{figure}[htbp]
\begin{promptbox}{System prompt}
\begin{verbatim}
You are a strict binary classifier for
Bengali hallucination detection. Given a
Bengali (context, prompt, response) triple,
you must output EXACTLY ONE character: the
digit 1 or the digit 0. No words, no JSON, no
explanation, no whitespace, no quotes — just
the digit.

OUTPUT 1  if the response is FAITHFUL: it is
supported by the context, or (when the
context is [NULL]) is factually correct.
OUTPUT 0  if the response is HALLUCINATED: it
contains factual errors, contradicts the
context, or invents information not
supported by it.
\end{verbatim}
\end{promptbox}
\vspace{5pt}
\begin{promptbox}{User template}
\begin{verbatim}
Context: {context}
Prompt: {prompt_bn}
Response: {response_bn}

Label (1 or 0 only):
\end{verbatim}
\end{promptbox}
\caption{The prompt of every zero-shot judge. Line breaks inside the system prompt are added for display.}
\label{fig:prompt}
\end{figure}

\para{Decoding.} Qwen2.5-32B, Gemma-2-9B and TigerLLM-9B-it run locally with Hugging Face Transformers in bfloat16,
decoding greedily for at most six new tokens. Llama-3.1-8B-Instruct runs through the OpenRouter API at temperature 0
with at most eight tokens, on providers that return log-probabilities. In both cases we keep the 20 most probable
tokens at the first generated position.

\para{Verdicts and scores.} The verdict is the first digit 0 or 1 in the output; if there is none, it is the score
thresholded at 0.5. The score is $s=P(1)/(P(1)+P(0))$, computed from the log-probabilities of the tokens \texttt{1}
and \texttt{0}; if only one of them is among the top 20, $s$ is $e^{\ell_1}$ or $1-e^{\ell_0}$, and if neither is,
$s=0.5$.

\para{Metrics.} Macro-F1 averages the per-class F1 over the classes present in the gold labels or the predictions, as
in scikit-learn; this is why a near-single-class subset can lose half its macro-F1 to one error
(\cref{app:single-class}). The false-flag rate divides faithful items judged hallucinated by all faithful items, and
the miss rate divides hallucinated items judged faithful by all hallucinated items. ROC-AUC is the probability that a
random faithful item scores above a random hallucinated one. ECE sorts items into 15 equal-width bins of the
confidence $\max(s,1-s)$ and averages $|\text{accuracy}-\text{confidence}|$ over the bins, weighted by bin size. The
released scorer, \texttt{scripts/evaluate.py}, reports all of these overall and per stratum.

\section{Additional Bengali-Adapted LLMs}
\label{app:bengali-llms}

Besides TigerLLM-9B-it (\cref{sec:baselines}), we tried two further Bengali-adapted generative models as zero-shot
judges, TituLLM-1B and BanglaLlama-7B. Both were evaluated with an earlier version of our prompt, which framed the model
as an expert Bengali fact-checker, asked whether the response was hallucinated, and inverted the verdict to the
dataset convention, so their numbers are not directly comparable to \cref{tab:headline}. \Cref{tab:app-bengali-llms}
reports them against the majority-class reference. TituLLM-1B is barely above that reference on either release, and
BanglaLlama-7B, run on a 50-item Phase-2 pilot only, labels every item as hallucinated (F1$_{\text{faith}}=0$).
Neither is usable as a judge at these scales, so both were dropped.

\begin{table}[htbp]
\centering
\small
\setlength{\tabcolsep}{4.2pt}
\begin{tabular}{@{}lrrrr@{}}
\toprule
\textbf{System} & \textbf{Acc.} & \textbf{Macro-F1} & \textbf{F1}$_{\text{hallu}}$ & \textbf{F1}$_{\text{faith}}$ \\
\midrule
\multicolumn{5}{@{}l}{\textit{Phase-1 test set (2{,}516 items)}} \\
Majority class & 0.531 & 0.347 & 0.000 & 0.693 \\
TituLLM-1B & 0.524 & 0.415 & 0.162 & 0.668 \\
\addlinespace
\multicolumn{5}{@{}l}{\textit{Phase~2 (5{,}000 items)}} \\
Majority class & 0.543 & 0.352 & 0.704 & 0.000 \\
TituLLM-1B & 0.467 & 0.400 & 0.201 & 0.600 \\
BanglaLlama-7B$^{\dagger}$ & 0.720 & 0.419 & 0.837 & 0.000 \\
\bottomrule
\end{tabular}

\caption{Two further Bengali-adapted LLMs as zero-shot judges beside the majority-class reference:
accuracy, macro-F1 and per-class F1 for hallucinated and faithful items. Both used an earlier prompt (see text), so
their scores are not comparable to \cref{tab:headline}. TituLLM-1B barely beats the reference, and BanglaLlama-7B labels
every item hallucinated (F1$_{\text{faith}}=0$). $^{\dagger}$50-item Phase-2 pilot only.}
\label{tab:app-bengali-llms}
\end{table}

\section{Teacher-Forced Log-Probability as a Hallucination Signal}
\label{app:logprob}

\begin{figure*}[tp]
\centering
\includegraphics[width=\textwidth]{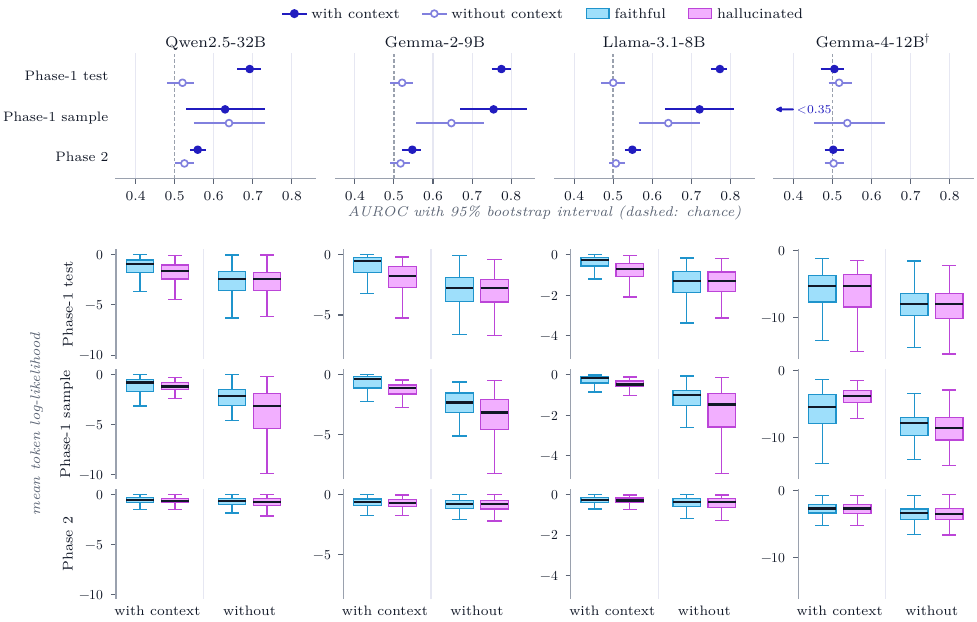}
\caption{Teacher-forced \field{mean\_logprob} as a hallucination signal, per model (columns). Top: AUROC with 95\%
bootstrap intervals on the Phase-1 test set, the released Phase-1 sample set and Phase~2; the arrow marks an AUROC
below the plotted range. Bottom: \field{mean\_logprob} of faithful and hallucinated responses, with and without a
context; boxes span the quartiles around the median, and whiskers reach the most extreme values within 1.5 times the
interquartile range. The sample set holds 130 items with a context (83 faithful, 47 hallucinated) and 169 without (80,
89). $^{\dagger}$Not one of our judges.}
\label{fig:logprob}
\end{figure*}

As a training-free complement to the verdict-classification judges of \cref{sec:baselines}, we teacher-force three of
them (all but TigerLLM-9B-it) on the gold \field{response\_bn}, appended to the same system, context and question
prompt, and record the mean per-token log-likelihood that each judge assigns to the actual response
(\field{mean\_logprob}), in the uncertainty-based tradition of hallucination detection. As a robustness check we also
include Gemma-4-12B \citep{gemmateam2026gemma4}, an open model that is not one of our judges. \Cref{tab:logprob}
reports how well \field{mean\_logprob} alone separates faithful from hallucinated responses (Cohen's $d$ and AUROC,
with 95\% bootstrap intervals over 1,000 resamples), separately for items with and without a supporting context, and
\cref{fig:logprob} adds the released sample set and the distributions behind every AUROC.

On the Phase-1 test set, Qwen2.5-32B, Gemma-2-9B and Llama-3.1-8B all separate the classes substantially when a
context is present (AUROC 0.69--0.78), but the signal falls to chance without one (AUROC 0.50--0.52), which suggests
that log-probability is informative only when there is evidence to contradict. The 299-item sample set is noisier:
the three judges reach 0.63--0.76 with a context and 0.64--0.65 without, with intervals that reach down to 0.53. On
Phase~2 the effect shrinks considerably (AUROC 0.51--0.56 with or without context), so teacher-forced log-probability
does not generalize across releases as a standalone detector. Gemma-4-12B shows essentially no separation on either
full release, and on the sample set's 130 context-grounded items its ranking is inverted (AUROC below 0.35).

\begin{table*}[tp]
\centering
\small
\setlength{\tabcolsep}{5pt}
\begin{tabular}{@{}lcccc@{}}
\toprule
 & \multicolumn{2}{c}{\textbf{With context}} & \multicolumn{2}{c}{\textbf{Without context}} \\
\cmidrule(lr){2-3}\cmidrule(l){4-5}
\textbf{Model} & \textbf{Cohen's} $d$ & \textbf{AUROC} & \textbf{Cohen's} $d$ & \textbf{AUROC} \\
\midrule
\multicolumn{5}{@{}l}{\textit{Phase-1 test set}\enspace{\color{muted}1{,}361 items with context, 1{,}155 without}} \\
Qwen2.5-32B & 0.489\,{\color{muted}[0.37, 0.62]} & 0.692\,{\color{muted}[0.66, 0.72]} & 0.051\,{\color{muted}[$-$0.05, 0.17]} & 0.520\,{\color{muted}[0.48, 0.55]} \\
Gemma-2-9B & 0.755\,{\color{muted}[0.65, 0.87]} & 0.775\,{\color{muted}[0.75, 0.80]} & 0.063\,{\color{muted}[$-$0.06, 0.18]} & 0.521\,{\color{muted}[0.49, 0.55]} \\
Llama-3.1-8B & 0.645\,{\color{muted}[0.53, 0.78]} & 0.773\,{\color{muted}[0.75, 0.79]} & 0.052\,{\color{muted}[$-$0.06, 0.15]} & 0.500\,{\color{muted}[0.47, 0.53]} \\
Gemma-4-12B$^{\dagger}$ & 0.031\,{\color{muted}[$-$0.08, 0.13]} & 0.505\,{\color{muted}[0.47, 0.53]} & 0.066\,{\color{muted}[$-$0.04, 0.18]} & 0.517\,{\color{muted}[0.49, 0.55]} \\
\addlinespace
\multicolumn{5}{@{}l}{\textit{Phase~2}\enspace{\color{muted}2{,}452 items with context, 2{,}548 without}} \\
Qwen2.5-32B & 0.201\,{\color{muted}[0.12, 0.28]} & 0.559\,{\color{muted}[0.54, 0.58]} & 0.153\,{\color{muted}[0.07, 0.23]} & 0.525\,{\color{muted}[0.50, 0.55]} \\
Gemma-2-9B & 0.149\,{\color{muted}[0.07, 0.23]} & 0.547\,{\color{muted}[0.52, 0.57]} & 0.114\,{\color{muted}[0.03, 0.18]} & 0.517\,{\color{muted}[0.49, 0.54]} \\
Llama-3.1-8B & 0.158\,{\color{muted}[0.08, 0.24]} & 0.549\,{\color{muted}[0.53, 0.57]} & 0.117\,{\color{muted}[0.04, 0.20]} & 0.507\,{\color{muted}[0.49, 0.53]} \\
Gemma-4-12B$^{\dagger}$ & 0.012\,{\color{muted}[$-$0.06, 0.10]} & 0.502\,{\color{muted}[0.48, 0.53]} & $-$0.022\,{\color{muted}[$-$0.10, 0.06]} & 0.503\,{\color{muted}[0.48, 0.53]} \\
\bottomrule
\end{tabular}

\caption{How well the teacher-forced mean log-likelihood of the response (\field{mean\_logprob}) alone
separates faithful from hallucinated responses: Cohen's $d$ (the standardized mean difference) and AUROC, with 95\%
bootstrap intervals in grey, on items with and without a supporting context. The signal is clear only with a context on
the Phase-1 test set (AUROC 0.69--0.78 for our three judges) and near chance elsewhere (0.50--0.56).
$^{\dagger}$Not one of our judges; included as a robustness check.}
\label{tab:logprob}
\end{table*}

\section{Failure-Mode Results}
\label{app:failure-modes}

\begin{figure*}[tp]
\centering
\includegraphics[width=\textwidth]{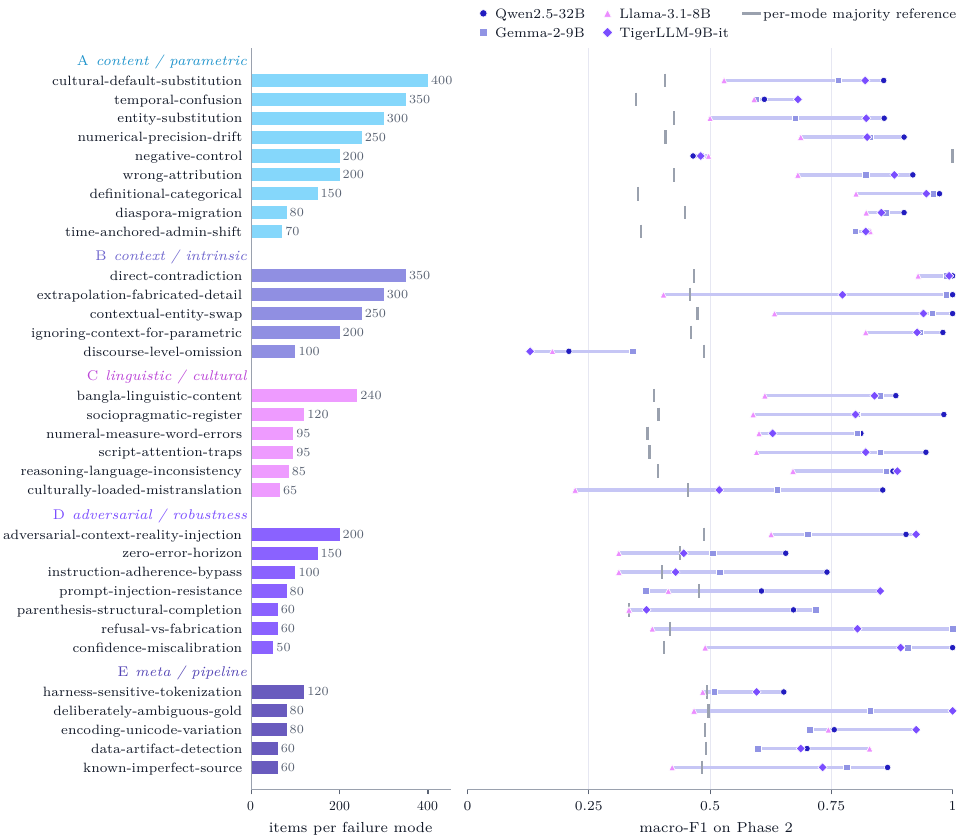}
\caption{Phase~2 by failure mode. Left: items per mode, colored by family. Right: each judge's macro-F1 on the mode
(markers), the spread across judges (band) and the per-mode majority reference (grey tick).}
\label{fig:modes}
\end{figure*}

\Cref{fig:modes} shows the composition of Phase~2 by failure mode beside every judge's macro-F1 on each mode, and
\cref{tab:failure-full} gives the values behind it, which also underlie \cref{tab:failure-value}.

\begin{table*}[tp]
\centering
\small
\setlength{\tabcolsep}{5pt}
\begin{tabular}{@{}clrrrrrr@{}}
\toprule
\textbf{Fam.} & \textbf{Failure mode} & $n$ & \textbf{Maj.} & \textbf{Qwen2.5} & \textbf{Gemma-2} & \textbf{Llama-3.1} & \textbf{TigerLLM} \\
\midrule
\multirow{9}{*}{A} & cultural-default-substitution & 400 & 0.407 & \textbf{0.858} & 0.764 & 0.529 & 0.820 \\
 & temporal-confusion & 350 & 0.348 & 0.612 & 0.595 & 0.591 & \textbf{0.681} \\
 & entity-substitution & 300 & 0.425 & \textbf{0.859} & 0.676 & 0.500 & 0.822 \\
 & numerical-precision-drift & 250 & 0.408 & \textbf{0.900} & 0.830 & 0.687 & 0.824 \\
 & negative-control & 200 & 1.000 & 0.465 & 0.481 & \textbf{0.497} & 0.481 \\
 & wrong-attribution & 200 & 0.425 & \textbf{0.918} & 0.821 & 0.681 & 0.880 \\
 & definitional-categorical & 150 & 0.351 & \textbf{0.973} & 0.960 & 0.801 & 0.946 \\
 & diaspora-migration & 80 & 0.448 & \textbf{0.900} & 0.862 & 0.822 & 0.853 \\
 & time-anchored-admin-shift & 70 & 0.358 & 0.827 & 0.799 & \textbf{0.830} & 0.821 \\
\midrule
\multirow{5}{*}{B} & direct-contradiction & 350 & 0.467 & \textbf{1.000} & 0.987 & 0.929 & 0.993 \\
 & extrapolation-fabricated-detail & 300 & 0.458 & \textbf{1.000} & 0.987 & 0.404 & 0.773 \\
 & contextual-entity-swap & 250 & 0.474 & \textbf{1.000} & 0.958 & 0.633 & 0.940 \\
 & ignoring-context-for-parametric & 200 & 0.461 & \textbf{0.980} & 0.934 & 0.821 & 0.927 \\
 & discourse-level-omission & 100 & 0.487 & 0.209 & \textbf{0.341} & 0.175 & 0.129 \\
\midrule
\multirow{6}{*}{C} & bangla-linguistic-content & 240 & 0.385 & \textbf{0.883} & 0.851 & 0.613 & 0.839 \\
 & sociopragmatic-register & 120 & 0.394 & \textbf{0.982} & 0.803 & 0.589 & 0.800 \\
 & numeral-measure-word-errors & 95 & 0.371 & \textbf{0.811} & 0.804 & 0.601 & 0.629 \\
 & script-attention-traps & 95 & 0.375 & \textbf{0.945} & 0.851 & 0.596 & 0.821 \\
 & reasoning-language-inconsistency & 85 & 0.393 & 0.877 & 0.863 & 0.671 & \textbf{0.886} \\
 & culturally-loaded-mistranslation & 65 & 0.454 & \textbf{0.856} & 0.639 & 0.222 & 0.519 \\
\midrule
\multirow{7}{*}{D} & adversarial-context-reality-injection & 200 & 0.488 & 0.904 & 0.702 & 0.626 & \textbf{0.925} \\
 & zero-error-horizon & 150 & 0.438 & \textbf{0.656} & 0.506 & 0.312 & 0.446 \\
 & instruction-adherence-bypass & 100 & 0.401 & \textbf{0.741} & 0.520 & 0.312 & 0.429 \\
 & prompt-injection-resistance & 80 & 0.477 & 0.606 & 0.368 & 0.414 & \textbf{0.851} \\
 & parenthesis-structural-completion & 60 & 0.333 & 0.672 & \textbf{0.718} & 0.333 & 0.369 \\
 & refusal-vs-fabrication & 60 & 0.417 & \textbf{1.000} & \textbf{1.000} & 0.381 & 0.804 \\
 & confidence-miscalibration & 50 & 0.405 & \textbf{1.000} & 0.908 & 0.490 & 0.893 \\
\midrule
\multirow{5}{*}{E} & harness-sensitive-tokenization & 120 & 0.494 & \textbf{0.652} & 0.509 & 0.485 & 0.596 \\
 & deliberately-ambiguous-gold & 80 & 0.497 & \textbf{1.000} & 0.830 & 0.467 & \textbf{1.000} \\
 & encoding-unicode-variation & 80 & 0.490 & 0.756 & 0.706 & 0.744 & \textbf{0.925} \\
 & data-artifact-detection & 60 & 0.492 & 0.700 & 0.598 & \textbf{0.829} & 0.687 \\
 & known-imperfect-source & 60 & 0.483 & \textbf{0.866} & 0.782 & 0.422 & 0.732 \\
\bottomrule
\end{tabular}

\caption{Macro-F1 of each zero-shot judge on each of the 32 Phase-2 failure modes, beside the mode's
majority reference (Maj., the macro-F1 of always predicting its majority label); $n$ is the mode's item count, and the
best judge per mode is in bold. Qwen2.5-32B is best on 22 modes, yet every judge stays below the reference on negative
controls and discourse-level omission.}
\label{tab:failure-full}
\end{table*}

\section{Single-Class Subsets and Error Counts}
\label{app:single-class}

Seven of the 32 Phase-2 failure modes are near-single-class, with at least 95\% of their items sharing one gold label,
and on such subsets macro-F1 can mislead. The clearest case is \mode{negative-control}: all 200 items are faithful, so
a system that never flags an item scores 1.000, while the first false flag introduces a hallucinated class with F1 of
0 and halves the score (\cref{tab:app-singleclass-example}). One error therefore costs 0.501 macro-F1 and the next 25
cost only 0.034. \Cref{tab:app-singleclass-modes} reports confusion counts, accuracy and both error rates for all
seven modes, and \cref{tab:app-error-rates-full} gives the false-flag and miss rates on every mode. Among the 32 ranked
Phase-2 teams, the false-flag rate on negative controls has a median of 8.2\% (mean 26.2\%), with 6 teams above 50\%;
their miss rate on discourse-level omission has a median of 59.5\%, with 19 of 32 teams above 50\%.

\begin{table*}[tp]
\centering
\small
\begin{minipage}{0.60\textwidth}
\centering
\begin{tabular}{rrr}
\toprule
\textbf{False flags (of 200)} & \textbf{Accuracy} & \textbf{Macro-F1} \\
\midrule
0 & 1.000 & 1.000 \\
1 & 0.995 & 0.499 \\
26 (Qwen2.5-32B) & 0.870 & 0.465 \\
\bottomrule
\end{tabular}

\caption{Accuracy and macro-F1 on the 200 negative-control items, all faithful, against
the number of false flags (faithful items judged hallucinated). The first false flag halves macro-F1 while accuracy
barely moves; the next 25, as in Qwen2.5-32B's verdicts, cost only 0.034 more.}
\label{tab:app-singleclass-example}
\end{minipage}

\vspace{12pt}
\setlength{\tabcolsep}{6pt}
\begin{tabular}{@{}llrrrrrrr@{}}
\toprule
 & & & & \multicolumn{2}{c}{\textbf{Counts}} & & \multicolumn{2}{c}{\textbf{Rates (\%)}} \\
\cmidrule(lr){5-6}\cmidrule(l){8-9}
\textbf{Failure mode} & \textbf{Judge} & $n$ & \textbf{\% faith.} & \textbf{FF} & \textbf{Miss} & \textbf{Acc.} & \textbf{FF} & \textbf{Miss} \\
\midrule
\multirow{4}{*}{negative-control} & Qwen & \multirow{4}{*}{200} & \multirow{4}{*}{100.0} & 26 & 0 & 0.870 & 13.0 & -- \\
 & Gemma &  &  & 15 & 0 & 0.925 & 7.5 & -- \\
 & Llama &  &  & 2 & 0 & 0.990 & 1.0 & -- \\
 & TigerLLM &  &  & 15 & 0 & 0.925 & 7.5 & -- \\
\midrule
\multirow{4}{*}{deliberately-ambiguous-gold} & Qwen & \multirow{4}{*}{80} & \multirow{4}{*}{98.8} & 0 & 0 & 1.000 & 0.0 & 0.0 \\
 & Gemma &  &  & 1 & 0 & 0.988 & 1.3 & 0.0 \\
 & Llama &  &  & 9 & 1 & 0.875 & 11.4 & 100.0 \\
 & TigerLLM &  &  & 0 & 0 & 1.000 & 0.0 & 0.0 \\
\midrule
\multirow{4}{*}{harness-sensitive-tokenization} & Qwen & \multirow{4}{*}{120} & \multirow{4}{*}{97.5} & 11 & 0 & 0.908 & 9.4 & 0.0 \\
 & Gemma &  &  & 16 & 2 & 0.850 & 13.7 & 66.7 \\
 & Llama &  &  & 4 & 3 & 0.942 & 3.4 & 100.0 \\
 & TigerLLM &  &  & 5 & 2 & 0.942 & 4.3 & 66.7 \\
\midrule
\multirow{4}{*}{data-artifact-detection} & Qwen & \multirow{4}{*}{60} & \multirow{4}{*}{96.7} & 5 & 0 & 0.917 & 8.6 & 0.0 \\
 & Gemma &  &  & 5 & 1 & 0.900 & 8.6 & 50.0 \\
 & Llama &  &  & 0 & 1 & 0.983 & 0.0 & 50.0 \\
 & TigerLLM &  &  & 2 & 1 & 0.950 & 3.4 & 50.0 \\
\midrule
\multirow{4}{*}{encoding-unicode-variation} & Qwen & \multirow{4}{*}{80} & \multirow{4}{*}{96.2} & 5 & 0 & 0.938 & 6.5 & 0.0 \\
 & Gemma &  &  & 4 & 1 & 0.938 & 5.2 & 33.3 \\
 & Llama &  &  & 0 & 2 & 0.975 & 0.0 & 66.7 \\
 & TigerLLM &  &  & 1 & 0 & 0.988 & 1.3 & 0.0 \\
\midrule
\multirow{4}{*}{adversarial-context-reality-injection} & Qwen & \multirow{4}{*}{200} & \multirow{4}{*}{95.5} & 4 & 0 & 0.980 & 2.1 & 0.0 \\
 & Gemma &  &  & 21 & 0 & 0.895 & 11.0 & 0.0 \\
 & Llama &  &  & 30 & 1 & 0.845 & 15.7 & 11.1 \\
 & TigerLLM &  &  & 3 & 0 & 0.985 & 1.6 & 0.0 \\
\midrule
\multirow{4}{*}{discourse-level-omission} & Qwen & \multirow{4}{*}{100} & \multirow{4}{*}{5.0} & 0 & 78 & 0.220 & 0.0 & 82.1 \\
 & Gemma &  &  & 0 & 60 & 0.400 & 0.0 & 63.2 \\
 & Llama &  &  & 0 & 82 & 0.180 & 0.0 & 86.3 \\
 & TigerLLM &  &  & 0 & 87 & 0.130 & 0.0 & 91.6 \\
\bottomrule
\end{tabular}

\caption{The seven near-single-class Phase-2 failure modes, on which at least 95\% of items
share one gold label: item count ($n$), faithful share (\% faith.) and, for each judge, its false flags (FF, faithful
items judged hallucinated) and misses (hallucinated items judged faithful) as counts, with its accuracy and error
rates; -- marks a class with no items. On such modes these counts, not macro-F1, show how a judge behaves.}
\label{tab:app-singleclass-modes}
\end{table*}

\begin{table*}[tp]
\centering
\small
\setlength{\tabcolsep}{4.5pt}
\begin{tabular}{@{}clrrrrrrrr@{}}
\toprule
 & & \multicolumn{2}{c}{\textbf{Qwen2.5}} & \multicolumn{2}{c}{\textbf{Gemma-2}} & \multicolumn{2}{c}{\textbf{Llama-3.1}} & \multicolumn{2}{c}{\textbf{TigerLLM}} \\
\cmidrule(lr){3-4}\cmidrule(lr){5-6}\cmidrule(lr){7-8}\cmidrule(l){9-10}
\textbf{Fam.} & \textbf{Failure mode} & FF & Miss & FF & Miss & FF & Miss & FF & Miss \\
\midrule
\multirow{9}{*}{A} & cultural-default-substitution & 12.7 & 12.8 & 4.0 & 31.8 & 1.6 & 67.5 & 1.6 & 24.5 \\
 & temporal-confusion & 52.8 & 24.1 & 43.6 & 37.4 & 4.9 & 66.8 & 13.5 & 47.1 \\
 & entity-substitution & 19.2 & 8.1 & 20.5 & 32.9 & 1.3 & 67.1 & 9.0 & 17.6 \\
 & numerical-precision-drift & 9.0 & 8.7 & 3.8 & 21.5 & 0.0 & 45.3 & 1.3 & 23.8 \\
 & negative-control & 13.0 & -- & 7.5 & -- & 1.0 & -- & 7.5 & -- \\
 & wrong-attribution & 7.7 & 6.1 & 7.7 & 18.2 & 0.0 & 41.9 & 5.8 & 11.5 \\
 & definitional-categorical & 1.2 & 4.3 & 1.2 & 7.2 & 1.2 & 39.1 & 1.2 & 10.1 \\
 & diaspora-migration & 4.6 & 13.3 & 1.5 & 33.3 & 0.0 & 46.7 & 0.0 & 40.0 \\
 & time-anchored-admin-shift & 17.9 & 16.1 & 20.5 & 19.4 & 0.0 & 35.5 & 7.7 & 29.0 \\
\midrule
\multirow{5}{*}{B} & direct-contradiction & 0.0 & 0.0 & 2.3 & 0.3 & 0.0 & 3.9 & 0.0 & 0.3 \\
 & extrapolation-fabricated-detail & 0.0 & 0.0 & 0.0 & 0.8 & 0.0 & 69.7 & 0.0 & 19.3 \\
 & contextual-entity-swap & 0.0 & 0.0 & 0.0 & 1.8 & 0.0 & 28.9 & 0.0 & 2.7 \\
 & ignoring-context-for-parametric & 0.0 & 1.2 & 3.4 & 3.5 & 6.9 & 11.7 & 0.0 & 4.7 \\
 & discourse-level-omission & 0.0 & 82.1 & 0.0 & 63.2 & 0.0 & 86.3 & 0.0 & 91.6 \\
\midrule
\multirow{6}{*}{C} & bangla-linguistic-content & 12.7 & 8.9 & 1.3 & 32.2 & 1.3 & 73.3 & 0.7 & 35.6 \\
 & sociopragmatic-register & 2.6 & 0.0 & 1.3 & 42.9 & 0.0 & 78.6 & 0.0 & 45.2 \\
 & numeral-measure-word-errors & 10.7 & 28.2 & 5.4 & 35.9 & 0.0 & 74.4 & 1.8 & 69.2 \\
 & script-attention-traps & 5.3 & 5.3 & 5.3 & 21.1 & 0.0 & 64.9 & 0.0 & 29.8 \\
 & reasoning-language-inconsistency & 26.7 & 1.8 & 10.0 & 14.5 & 3.3 & 49.1 & 10.0 & 10.9 \\
 & culturally-loaded-mistranslation & 9.1 & 9.3 & 9.1 & 35.2 & 0.0 & 92.6 & 0.0 & 55.6 \\
\midrule
\multirow{7}{*}{D} & adversarial-context-reality-injection & 2.1 & 0.0 & 11.0 & 0.0 & 15.7 & 11.1 & 1.6 & 0.0 \\
 & zero-error-horizon & 39.3 & 6.1 & 35.0 & 60.6 & 41.9 & 100.0 & 32.5 & 78.8 \\
 & instruction-adherence-bypass & 15.2 & 29.9 & 18.2 & 62.7 & 0.0 & 94.0 & 12.1 & 77.6 \\
 & prompt-injection-resistance & 30.1 & 0.0 & 65.8 & 0.0 & 32.9 & 85.7 & 6.8 & 0.0 \\
 & parenthesis-structural-completion & 13.3 & 50.0 & 3.3 & 50.0 & 0.0 & 100.0 & 0.0 & 96.7 \\
 & refusal-vs-fabrication & 0.0 & 0.0 & 0.0 & 0.0 & 14.0 & 100.0 & 0.0 & 47.1 \\
 & confidence-miscalibration & 0.0 & 0.0 & 12.5 & 5.9 & 0.0 & 73.5 & 0.0 & 14.7 \\
\midrule
\multirow{5}{*}{E} & harness-sensitive-tokenization & 9.4 & 0.0 & 13.7 & 66.7 & 3.4 & 100.0 & 4.3 & 66.7 \\
 & deliberately-ambiguous-gold & 0.0 & 0.0 & 1.3 & 0.0 & 11.4 & 100.0 & 0.0 & 0.0 \\
 & encoding-unicode-variation & 6.5 & 0.0 & 5.2 & 33.3 & 0.0 & 66.7 & 1.3 & 0.0 \\
 & data-artifact-detection & 8.6 & 0.0 & 8.6 & 50.0 & 0.0 & 50.0 & 3.4 & 50.0 \\
 & known-imperfect-source & 1.8 & 25.0 & 5.4 & 25.0 & 42.9 & 50.0 & 3.6 & 50.0 \\
\bottomrule
\end{tabular}

\caption{False-flag rate (FF, the share of faithful items judged hallucinated) and miss rate
(the share of hallucinated items judged faithful), in \%, of each judge on all 32 Phase-2 failure modes; -- marks a
class with no items. Llama-3.1-8B rarely flags a faithful item but misses most hallucinations (median miss rate
67\%), whereas Qwen2.5-32B keeps both rates low (medians 8.2\% and 5.3\%).}
\label{tab:app-error-rates-full}
\end{table*}

\FloatBarrier
\section{Competition Details}
\label{app:competition}

\Cref{tab:comp-tiers} bins the teams of both phases by their best macro-F1, and \cref{tab:overlap-full} lists every
team that competed in both phases (\cref{fig:phases} plots them). Team names are as they appear on the public
leaderboard.

\begin{table}[htbp]
\centering
\small
\begin{tabular}{lrr}
\toprule
\textbf{Best macro-F1} & \textbf{Phase-1 teams} & \textbf{Phase-2 teams} \\
\midrule
$<0.60$ & 18 & 6 \\
$0.60$--$0.70$ & 53 & 2 \\
$0.70$--$0.80$ & 77 & 5 \\
$0.80$--$0.90$ & 42 & 12 \\
$\geq 0.90$ & 32 & 7 \\
\midrule
Total teams & 222 & 32 \\
\bottomrule
\end{tabular}

\caption{Teams by their best private-leaderboard macro-F1, in bins, in Phase~1 (222 teams) and Phase~2
(32 ranked teams). Most Phase-1 teams score 0.70--0.90, while the Phase-2 field spreads wider, with 6 teams below 0.60
and 7 at 0.90 or above.}
\label{tab:comp-tiers}
\end{table}

\begin{table}[htbp]
\centering
\small
\begin{tabular}{@{}lrrr@{}}
\toprule
\textbf{Team} & \textbf{Phase 1} & \textbf{Phase 2} & \textbf{Change} \\
\midrule
ExtremeAI & 0.983 & 0.812 & \textcolor{loss}{$-$0.172} \\
Huntrix & 0.983 & 0.901 & \textcolor{loss}{$-$0.081} \\
BUET\_NeuralX & 0.962 & 0.954 & \textcolor{loss}{$-$0.008} \\
IUFree & 0.959 & 0.891 & \textcolor{loss}{$-$0.068} \\
Ain't No Way & 0.959 & 0.381 & \textcolor{loss}{$-$0.578} \\
Bitwiz & 0.936 & 0.921 & \textcolor{loss}{$-$0.015} \\
DataDriven & 0.933 & 0.603 & \textcolor{loss}{$-$0.330} \\
sudo\_rm\_-rf & 0.930 & 0.710 & \textcolor{loss}{$-$0.221} \\
r=0 & 0.928 & 0.786 & \textcolor{loss}{$-$0.142} \\
ArektaGenericTeam & 0.927 & 0.833 & \textcolor{loss}{$-$0.094} \\
CUET cURL-marx & 0.926 & 0.673 & \textcolor{loss}{$-$0.253} \\
4ier & 0.923 & 0.905 & \textcolor{loss}{$-$0.018} \\
CUET\_AL\_Masaar & 0.923 & 0.830 & \textcolor{loss}{$-$0.093} \\
zero\_shot & 0.919 & 0.822 & \textcolor{loss}{$-$0.097} \\
CutiePookieUrza & 0.912 & 0.898 & \textcolor{loss}{$-$0.014} \\
Reverse\_Attention & 0.912 & 0.887 & \textcolor{loss}{$-$0.024} \\
watermelon & 0.910 & 0.704 & \textcolor{loss}{$-$0.206} \\
The Hallucinators & 0.902 & 0.749 & \textcolor{loss}{$-$0.152} \\
wrongCha & 0.898 & 0.897 & \textcolor{loss}{$-$0.000} \\
The Hunting Party & 0.897 & 0.370 & \textcolor{loss}{$-$0.527} \\
Labyrinth & 0.886 & 0.805 & \textcolor{loss}{$-$0.082} \\
SUST\_Nabajatra & 0.874 & 0.477 & \textcolor{loss}{$-$0.397} \\
DU\_Outliers & 0.872 & 0.902 & \textcolor{gain}{$+$0.030} \\
GTA8 & 0.837 & 0.429 & \textcolor{loss}{$-$0.408} \\
byte\_the\_dust & 0.824 & 0.371 & \textcolor{loss}{$-$0.453} \\
\bottomrule
\end{tabular}

\caption{All 25 teams that competed in both phases: best private-leaderboard macro-F1 on Phase~1 (2,516
items) and Phase~2 (5,000 items), sorted by Phase-1 score, and the change (Phase~2 minus Phase~1; losses in red). Only
DU\_Outliers improved, and the Phase-1 winner, ExtremeAI, fell from 0.983 to 0.812.}
\label{tab:overlap-full}
\end{table}

\end{document}